\documentclass[sigconf,screen]{acmart}
\AtBeginDocument{%
  \providecommand\BibTeX{{%
    \normalfont B\kern-0.5em{\scshape i\kern-0.25em b}\kern-0.8em\TeX}}}

\setcopyright{acmlicensed}
\copyrightyear{2018}
\acmYear{2018}
\acmDOI{XXXXXXX.XXXXXXX}

\acmConference[Pre-print]{}{Under Review}{}

\acmISBN{978-1-4503-XXXX-X/18/06}

\usepackage{microtype}
\usepackage{graphicx}
\usepackage{subfigure}
\usepackage{booktabs}
\usepackage{hyperref}

\usepackage{amsmath}
\usepackage{mathtools}
\usepackage{amsthm}
\usepackage[capitalize,noabbrev]{cleveref}
\usepackage{balance}

\theoremstyle{plain}
\newtheorem{theorem}{Theorem}[section]

\newtheorem{postulate}[theorem]{Postulate}

\theoremstyle{definition}

\theoremstyle{remark}

\usepackage[textsize=tiny]{todonotes}

\usepackage{amsmath,bm}

\def\eqref#1{equation~\ref{#1}}

\def\1{\bm{1}}

\DeclareMathAlphabet{\mathsfit}{\encodingdefault}{\sfdefault}{m}{sl}
\SetMathAlphabet{\mathsfit}{bold}{\encodingdefault}{\sfdefault}{bx}{n}

\DeclareMathOperator*{\argmin}{arg\,min}

\usepackage{url}
\usepackage{algorithm}
\usepackage{listings}
\usepackage{xcolor}
\usepackage{dsfont}
\usepackage{colortbl}
\usepackage{soul}
\definecolor{codeblue}{rgb}{0,0.4,0.6}
\lstdefinestyle{mystyle}{
    commentstyle=\color{codeblue},
    basicstyle=\ttfamily\footnotesize,
    breakatwhitespace=false,         
    breaklines=true,                 
    captionpos=b,                    
    keepspaces=true,                 
    numbers=left,                    
    numbersep=5pt,  
    frame=none,
    showspaces=false,                
    showstringspaces=false,
    showtabs=false,                  
    literate={\ \ }{{\ }}1,
    basicstyle=\fontsize{9}{13}\selectfont\ttfamily
}

\usepackage{adjustbox}
\usepackage{booktabs}
\usepackage{multirow}
\usepackage{multicol}
\usepackage{tikz}
\usepackage{pgfplots}
\usepackage{amsmath}
\usepackage{amsfonts}
\usepackage{mathtools}
\usepackage{newfloat}
\usepackage{listings}
\usepackage{enumitem}
\usepackage{csquotes}
\pgfplotsset{compat=1.8}

\floatstyle{ruled}
\newfloat{listing}{tb}{lst}{}
\floatname{listing}{Listing}

\usepackage{xargs}

\NewDocumentCommand{\reals}{}{\mathds{R}}

\NewDocumentCommand{\indsymb}{}{\mathds{1}}
\DeclareMathOperator{\expectsymb}{\mathds{E}}

\DeclareMathOperator{\probsymb}{\mathds{P}}

\newcommand{\featurespace}{\mathcal{X}}

\newcommand{\tokenspace}{\mathcal{V}}
\newcommand{\numlabels}{L}

\newcommand{\featureextract}{\Phi}
\newcommand{\surrogatemodel}{\Psi}
\newcommand{\embspace}{\mathcal{E}}

\NewDocumentCommand{\wrapbrackets}{m}{\left[ #1 \right]}

\NewDocumentCommand{\expectation}{e{_}o}{
    \def\decorated{\expectsymb\IfNoValueTF{#1}{}{_{#1}}}
    \IfNoValueTF{#2}{\decorated}{\decorated\!\wrapbrackets{#2}}
}

\NewDocumentCommand{\probability}{o}{
    \IfNoValueTF{#1}{\probsymb}{\probsymb\!\wrapbrackets{#1}}
}

\NewDocumentCommand{\indicator}{o}{
    \IfNoValueTF{#1}{\indsymb}{\indsymb\!\wrapbrackets{#1}}
}

\NewDocumentCommand{\intrange}{o}{
    \wrapbrackets{#1}
}

 \NewDocumentCommand{\set}{m}{
    \left\{#1\right\}
 }

\NewDocumentCommand{\labelvec}{o}{
    \IfNoValueTF{#1}{\mathbf{y}}{\mathbf{y}_{#1}}
}

\NewDocumentCommand{\labeldec}{o}{
    \IfNoValueTF{#1}{\mathbf{W}}{\mathbf{w}_{#1}}
}

\NewDocumentCommand{\labelfeature}{o}{
    \IfNoValueTF{#1}{\mathbf{z}}{\mathbf{z}_{#1}}
}

\NewDocumentCommand{\classifier}{o}{
    \IfNoValueTF{#1}{\mathcal{W}}{\mathcal{W}_{#1}}
}
\NewDocumentCommand{\instance}{o}{
\IfNoValueTF{#1}{\mathbf{x}}{\mathbf{x}_{#1}}
}

\begin{document}

\title{Learning label-label correlations in \\ Extreme Multi-label Classification via Label Features}
\author{Siddhant Kharbanda}
\affiliation{\institution{University of California, Los Angeles} \city{} \country{}}
\author{Devaansh Gupta}
\affiliation{\institution{Aalto University} \city{} \country{}}
\author{Erik Schultheis}
\affiliation{\institution{Aalto University} \city{} \country{}}
\author{Atmadeep Banerjee}
\affiliation{\institution{Aalto University} \city{} \country{}}
\author{Cho-Jui Hsieh}
\affiliation{\institution{University of California, Los Angeles} \city{} \country{}}
\author{Rohit Babbar}
\affiliation{\institution{University of Bath \\ Aalto University} \city{} \country{}}
\renewcommand{\shortauthors}{Kharbanda, et al.}

\begin{abstract}
  Extreme Multi-label Text Classification (XMC) involves learning a classifier that can assign an input with a subset of most relevant labels from millions of label choices. Recent works in this domain have increasingly focused on a symmetric problem setting where both input instances and label features are short-text in nature. Short-text XMC with label features has found numerous applications in areas such as query-to-ad-phrase matching in search ads, title-based product recommendation, prediction of related searches. In this paper, we propose \textit{Gandalf}, a novel approach which makes use of a label co-occurrence graph to leverage label features as additional data points to supplement the training distribution. By exploiting the characteristics of the short-text XMC problem, it leverages the label features to construct valid training instances, and uses the label graph for generating the corresponding soft-label targets, hence effectively capturing the label-label correlations. Surprisingly, models trained on these new training instances, although being less than half of the original dataset, can outperform models trained on the original dataset, particularly on the PSP@k metric for tail labels. With this insight, we aim to train existing XMC algorithms on both, the original and new training instances, leading to an average 5\% relative improvements for 6 state-of-the-art algorithms across 4 benchmark datasets consisting of up to 1.3M labels. \textsc{Gandalf} can be applied in a plug-and-play manner to various methods and thus forwards the state-of-the-art in the domain, without incurring any additional computational overheads.
\end{abstract}
\maketitle

\section{Introduction}
\emph{Extreme Multilabel Classification} (XMC) has found numerous applications in the domains of related searches~\citep{Jain2019Slice}, dynamic search advertising~\citep{Parabel} and recommendation tasks,
which require predicting the most relevant results that frequently co-occur together \citep{clusterGCN, OpenGraphBenchmark}, or are highly correlated to the given product or search query.
These tasks are often modeled through embedding-based retrieval-cum-ranking pipelines over millions of possible web page titles, products titles, or ad-phrase keywords forming the label space.

Going beyond conventional tagging tasks for long textual documents consisting of hundreds of words, such as articles in encyclopedia \citep{partalas2015lshtc}, and bio-medicine \citep{tsatsaronis2015overview}, contemporary research focus has also widened to settings in which the input is just a short phrase, such as a search query or product title. Propelled by the surge in online search, recommendation, and advertising, applications of short-text XMC ranging from query-to-ad-phrase prediction~\citep{Astec} to title-based product-to-product~\citep{Decaf} recommendation have become increasingly prominent.

A major challenge across XMC problems is the extreme imbalance observed in their data distribution. Specifically, these datasets adhere to Zipf's
law~\citep{adamic2002zipf,APLC_XLNet}, i.e., following a long-tailed
distribution, where most labels are tail labels with very few ($\leq 5$)
positive data-points in a training set spanning $\geq 10^6$ total data points (\autoref{tab:datasets}). With so few positive examples, training a successful classifier on these labels purely from instance-to-label pairs seems an
insurmountable challenge. Therefore, recent methods have begun to incorporate additional data sources.

\paragraph{\textbf{Label features and label co-occurrence}}
In many of the settings listed above, labels are not just featureless integers, but do have a semantic meaning in and of themselves. For example, when matching products, each product ID could be associated with the name of the product. This is particularly attractive in the short-text setting, when both inputs and labels come from the \emph{same} space of short phrases. Consequently, while earlier work mostly focused on the nuances of short-text inputs \citep{Astec, kharbanda2021embedding}, more recent methods have successfully incorporated the short-text label descriptors into their pipeline~\citep{Decaf, Eclare, Siamese, NGAME}.

Yet, this still seems to underutilize the wealth of information present in label features. In particular, we demonstrate that it is possible to train a classifier using \emph{only} label information, that is, without ever presenting to it any of the training instances, and \emph{outperform} the same classifier trained on the original training data on tail labels. This surprising feat is enabled by the exploitation of label co-occurrence information.

In particular, using the interchangability of label features and instances, instead of aiming for contrastive learning~\citep{Siamese}, we want to use the label features as additional, supervised training points. However, this requires them to be associated with some apriori unknown label vector. In order to generate training targets, we make the assumption that the probability of a label $j$ being relevant for the textual feature of another label $i$, is equal to the conditional probability of observing $j$, given that $i$ is also a relevant label.

\paragraph{\textbf{Contributions}}
This insight yields a simple method, \emph{Gandalf}
(\textbf{G}raph \textbf{A}ugme\textbf{N}ted \textbf{DA}ta with \textbf{L}abel \textbf{F}eatures),
which exploits the unique setting of short-text XMC in a novel manner to generate additional training data in order to alleviate the data scarcity problem. As a data-centric approach, it is independent of the specific model architecture, enabling its application to a wide range of both current and potential future state-of-the-art models. The unchanged model architecture also implies that not only the model inference latency remains unchanged, but also peak memory consumption required during training is unaffected, contrary to some model-based approaches that
incorporate label metadata~\citep{Decaf,Siamese,chien2023pina}.

The additional training instances lead to overall longer training time.
Nonetheless, when keeping the compute budget fixed, we can observe
\emph{Gandalf} significantly outperforming the original dataset. When trained until convergence, we show an average of 5\% improvement on 5 state-of-the-art extreme classifiers across 4 public short-text benchmarks, with some settings seeing gains up to 30\%. In this way, XMC methods which inherently do not leverage label features can beat or perform on par with strong baselines which either employ elaborate training pipelines \citep{Siamese}, large transformer encoders~\citep{attentionxml, zhang2021fast, NGAME} or make heavy architectural modifications~\citep{Decaf, Eclare} to leverage label features.

Finally, we show that Gandalf could be considered an extension of the GLaS~\citep{Guo2019} regularizer to the label feature setting. We interpret it as tuning the bias-variance trade-off, where the additional error introduced by inaccurate additional training data is more then compensated for by the decrease is variance, especially for extremely noise tail labels~\citep{anonymous2024enhancing}.

\section{Preliminaries}
\label{sec:preliminaries}
For training, we have available a multi-label dataset $\mathcal{D} = \\ \left(\{\instance[i], \labelvec[i]\}^N_{i=1}, \{\labelfeature[l]\}_{l=1}^L\right)$
comprising of $N$ data points. Each $i \in \intrange[N]$ is associated with a small ground truth label vector $\labelvec[i] \in \set{0,1}^{\smash{\numlabels}}$ from $L \sim 10^6$ possible labels. Further, $\instance[i], \labelfeature[l] \in \featurespace$ denote the textual descriptions of the data point $i$ and the label $l$ which, in this setting, derive from the same vocabulary universe $\tokenspace$ \citep{Siamese}.
The goal is to learn a parameterized function $f\colon \instance[i] \mapsto \labelvec[i]$.

\paragraph{\textbf{One-vs-All Classification (OvA)}}
A common strategy for handling this learning problem is to map instances and labels
into a common Euclidean space $\embspace = \reals^{\smash{d}}$, 
in which the relevance $s_l(\instance)$ of a label $l$ to an instance is scored using 
an inner product, $s_l(\instance) = \langle \featureextract(\instance),
\labeldec[l] \rangle$. Here, $\featureextract(\instance)$ is the embedding of
instance $\instance$, and $\labeldec[l]$ the $l$'th column of the weight matrix
$\labeldec$.

The prediction function selects the $k$ highest-scoring labels, $f(\instance) = \operatorname{top}_k \left( \langle \featureextract(\instance), \labeldec \rangle \right)$. Training is usually handled using the \emph{one-vs-all} paradigm,
which applies a binary loss function $\ell$ to each entry in the score vector. In practice, performing the sum over all labels for each instance is 
prohibitively expensive, so the sum is approximated by a shortlist of labels $\mathcal{S}(\instance[i])$ that typically
contains all the positive labels, and only those negative labels which are expected to be challenging for classification
\citep{Siamese, Astec, NGAME, zhang2021fast, kharbanda2021embedding}:
\begin{equation}
    \begin{split}
        \mathcal{L}_{\mathcal{D}}[\featureextract, \labeldec] = \sum_{i=1}^{N} \sum_{l=1}^{\numlabels} \ell(\labelvec[il], \langle \featureextract(\instance), \labeldec[l] \rangle) \\ \approx \sum_{i=1}^{N} \sum_{l \in \mathcal{S}(\instance[i])} \ell(\labelvec[il], \langle \featureextract(\instance),
        \labeldec[l] \rangle)  \,.
    \end{split} \label{eq:xmc-task}
\end{equation}

Even though these approaches have been used with success, they still
struggle in learning good embeddings $\labeldec[l]$ for tail labels: A
classifier that learns solely based on instance-label pairs has little chance of
learning similar label representations for labels that do not co-occur within
the dataset, even though they might be semantically related. Consequently, 
training can easily lead to overfitting even with simple classifiers \citep{Guo2019}.

\begin{table}[!t]
    \centering
    \begin{adjustbox}{width = \columnwidth, center}
    \begin{tabular}{cccccc}
        \toprule
        \textbf{Datasets} & \textbf{N} & \textbf{L} & \textbf{APpL} & \textbf{ALpP} & \textbf{AWpP} \\
        \midrule
        LF-AmazonTitles-131K & 294,805 & 131,073 & 5.15 & 2.29 & 6.92\\
        LF-WikiSeeAlsoTitles-320K & 693,082 & 312,330 & 4.67 & 2.11 & 3.01\\
        LF-WikiTitles-500K  & 1,813,391 & 501,070 & 17.15 & 4.74 & 3.10\\
        LF-AmazonTitles-1.3M & 2,248,619 & 1,305,265 & 38.24 & 22.20 & 8.74\\
        \bottomrule
    \end{tabular}
    \end{adjustbox}
    \caption{Details of short-text benchmarks with label features. APpL is the avg. points per label, ALpP being avg. labels per point and AWpP is the length i.e. avg. words per point.}
    \label{tab:datasets}
    \vspace{-1em}
\end{table}

\paragraph{\textbf{Label Features}}
To reduce the generalization gap, regularization needs to be applied to the
label weights $\labeldec$, either explicitly as a new term in the loss
function \citep{Guo2019}, or implicitly through the inductive biases of the
network structure \citep{Decaf, Eclare} or by a learning algorithm \citep{Siamese, NGAME}. 
These approaches incorporate additional label metadata -- \emph{label features} -- to generate the inductive biases.
For short-text XMC, these features themselves are often short textual description, coming from the same space as the instances, as the following examples, taken from (i) LF-AmazonTitles-131K (recommend related products given a product name)
and (ii) LF-WikiTitles-500K (predict relevant categories, given the title of a Wikipedia page) illustrate:

\begin{figure*}[!t]
    \centering
    \includegraphics[width=\textwidth]{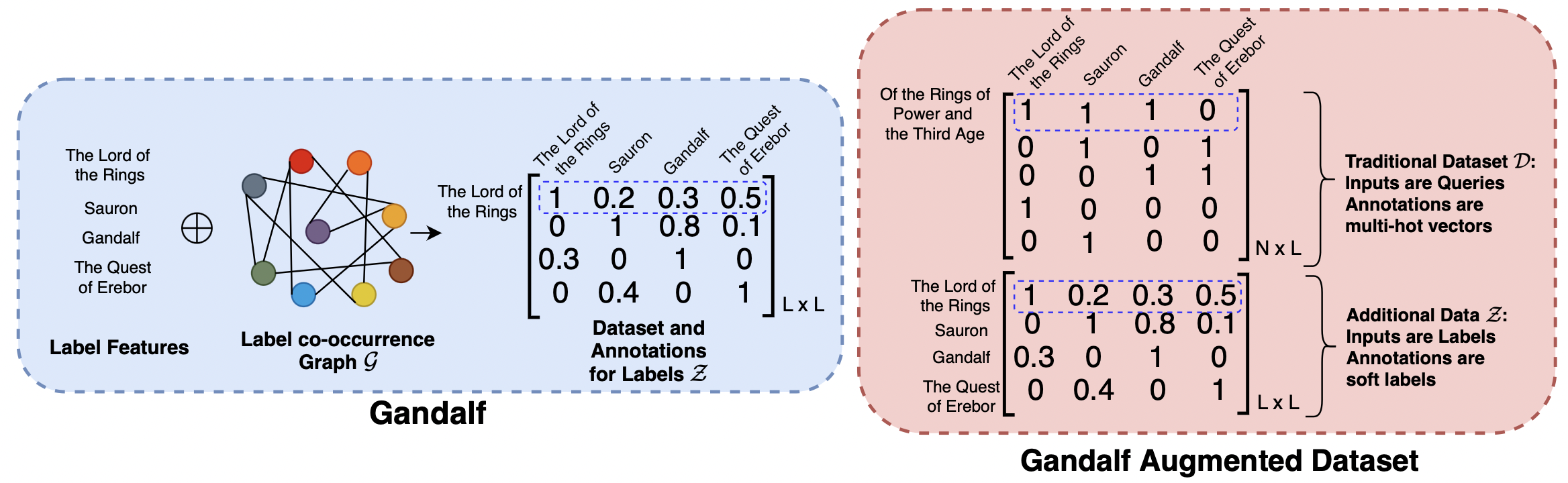}
    \caption{
    Gandalf augments the training dataset $\mathcal{D}$ by generating  soft
    targets for each label based on label co-occurrence statistics. 
    These additional datapoints $\mathcal{Z}$ are simply concatenated to the traditional
    dataset for training.}
    \label{fig:Gandalf}
\end{figure*}

\noindent \emph{Example 1:} For \underline{\textit{``Mario Kart: Double Dash!!''}} on Amazon, we have available: \textit{Mario Party 7} $\vert$ \textit{Super Smash Bros Melee} $\vert$ \textit{Super Mario Sunshine} $\vert$ \textit{Super Mario Strikers} as the recommended products. 

\noindent \emph{Example~2:} For the \underline{\textit{``2022 French presidential election''}} Wikipedia page, we have the available categories: \textit{April 2022 events in France} $\vert$ \textit{2022 French presidential election} $\vert$ \textit{2022 elections in France} $\vert$ \textit{Presidential elections in France}. Further, a google search of the same query, leads to the following related searches - \textit{French election 2022 - The Economist} $\vert$ \textit{French presidential election coverage on FRANCE 24} $\vert$ \textit{Presidential Election 2022: A Euroclash Between a “Liberal...} $\vert$  \textit{French polls, trends and election news for France}, amongst others. 

In view of these examples, one can affirm two important observations:
(i) the short-text XMC problem indeed requires recommending similar items which are either highly correlated or co-occur frequently with the queried item, and
(ii) the queried item and the corresponding label-features form an ``equivalence class'' and convey similar intent \citep{Siamese}.
For example, a valid news headline search should either result in a page mentioning the same headline or similar headlines from other media outlets (see Example 2).
As a result, it can be argued that data instances are \emph{interchangeable} with their respective labels' features.

Exploiting this interchangeability of label and instance text, \citet{Siamese, NGAME} proposes to tie encoder and decoder together and require $\labeldec[l] = \featureextract(\labelfeature[l])$.
While indeed yielding improved test performance, the condition $\labeldec[l] = \featureextract(\labelfeature[l])$ turns out to be too strong, and it has to allow for some fine-tuning corrections $\boldsymbol \eta_l$, yielding
$\labeldec[l] = \featureextract(\labelfeature[l]) + \boldsymbol \eta_l$.
Consequently, training of \textsc{SiameseXML} and \textsc{NGAME} is done in two stages: a contrastive loss needs to be minimized, followed by
fine-tuning with a classification objective.

\paragraph{\textbf{Label correlations}}
Label-label dependencies can appear in multi-label classification in two
different forms: \emph{Conditional} label correlations, and \emph{marginal} label correlations~\citep{dembczynski2012label}. In the conditional case, label dependencies are considered conditioned on each individual query, that is, they are independent if\footnote{Capital $X$ and $\mathbf{Y}$ denote the random
variables associated with instance and labels, rsp.}
\begin{equation}
    \probability[\mathbf{Y} \mid X] = {\textstyle\prod_j} \probability[Y_j \mid X] \,.
\end{equation}
As an example,
consider the search query \textit{``Jaguar''}: If we know just this search term, the results pertaining to both, the car brand and the animal, are likely to be relevant. However, knowing that during a particular instance of this search, the user was interested in the animal, one can conclude that car-based labels are less likely to be relevant. In this way, the presence of one label gives information \emph{beyond} what can be extracted just from the search query.

On the other hand, similar labels will generally appear together. Taking example 2 from the previous section, labels \textit{``2022 events in France''} and \textit{``2022 elections in France''} will have an above-random chance of
occurring together; however, that information is already carried in the query \textit{``2022 French presidential election''}, so the presence of one of these labels doesn't provide any new information, \emph{given} the query. In that sense, labels are marginally independent if
\begin{equation}
    \probability[\mathbf{Y}] = {\textstyle\prod_j} \probability[Y_j] \,.
\end{equation}

Given an instance, OvA classifiers generate scores independently for all labels. Thus, they are \emph{fundamentally incapable} of modelling
conditional label dependence. However, as standard performance metrics (P@k, PSP@k) are also decomposable into independent contributions of each label, that is, they can be expressed purely in terms of label marginals, they are similarly incapable of detecting whether a classifier models conditional label dependence~\citep{dembczynski2012label}.

This means that, as long as we want to focus on these standard metrics (and not on inter-dependency aware losses such as \citet{hullermeier2022flexible}), we only need to care about marginal correlations. At first glance, this seems trivial: It can be shown that an OvA classifier, trained using a proper loss, is consistent for P@k~\citep{Wydmuch18,menon2019multilabel}. Unfortunately, consistency only tells us that, in the limit of infinite training data, we will get a Bayes-optimal classifier. However, in practice, the XMC setting is very far from infinite data---most tail labels will have less than five positive training examples.

Thus, the question we want to tackle here is: Can we exploit knowledge about marginal label correlations to improve training in the data-scarce regime of long-tailed multi-label problems?

\section{Gandalf: Learning From Label-Label Correlations}
\label{sec:gandalf}

By combining marginal label correlations with label features, we can 
extend the \emph{self-annotation postulate} of \citet{Siamese} to:
\begin{postulate}{Label-feature Annotation:}
\label{prop:soft-self-annot}
    Given a label $j$ with label-features $\labelfeature[j]$, we posit that if these
    features are posed as a data point, its labels should follow the marginal label correlations, that is
    \begin{equation}
        \probability[Y_i = 1 \mid X = \labelfeature[j]] \approx \probability[Y_i = 1 \mid Y_j=1] \,.
        \label{eq:label-feature-annotation}
    \end{equation}
\end{postulate}
Note that this reduces to self-annotation by setting $i=j$, in which case
\eqref{eq:label-feature-annotation} becomes $\probability[Y_j \mid
\labelfeature[j]] \approx \probability[Y_j \mid Y_j] = 1$.

\begin{table*}[!h]
\begin{adjustbox}{width=\textwidth,center}
\begin{tabular}{l|ccc|ccc|ccc|ccc}
\toprule
& \multicolumn{6}{c|}{\textbf{LF-AmazonTitles-131K}} & \multicolumn{6}{c}{\textbf{LF-WikiSeeAlsoTitles-320K}} \\
\specialrule{0.70pt}{0.4ex}{0.65ex}
\textbf{Training Data} & \textbf{P@1}   & \textbf{P@3}   & \textbf{P@5}   & \textbf{PSP@1} & \textbf{PSP@3} & \textbf{PSP@5} & \textbf{P@1}   & \textbf{P@3}   & \textbf{P@5}   & \textbf{PSP@1} & \textbf{PSP@3} & \textbf{PSP@5} \\
\midrule
original $\mathcal{D}$ & 35.62 & 24.13 & 17.35 & 27.53 & 33.06 & 37.50 & 21.53 & 14.19 & 10.66 & 13.06 & 14.87 & 16.33 \\
surrogate $\mathcal{Z}$ & 29.68 & 21.47 & 16.04 & 28.76 & 33.75 & 38.27 & 22.88 & 16.02 & 12.44 & 22.03 & 23.69 & 25.55 \\
$\mathcal{G} = \mathcal{Z} \cup \mathcal{D}$ & \textbf{43.52} & \textbf{29.23} & \textbf{20.92} & \textbf{36.96} & \textbf{42.71} & \textbf{47.64}   & \textbf{31.31} & \textbf{21.38} & \textbf{16.22} & \textbf{24.31} & \textbf{26.79} & \textbf{28.83}\\
$\mathcal{U}(\mathcal{G}, N)$ & 38.46 & 25.81 & 18.52 & 32.29 & 37.17 & 41.59 & 25.93 & 17.54 & 13.34 & 19.75 & 21.76 & 23.57 \\
$\mathcal{Z}^1 \cup \mathcal{D}$ & 37.59 & 25.25 & 18.18 & 30.75 & 35.54 & 40.06  & 24.43 & 16.16 & 12.15 & 16.89 & 18.45 & 20.02\\
\bottomrule
\end{tabular}
\end{adjustbox}
\caption{Experiments showing the quality of the datasets created with label features on InceptionXML. While the baseline is surpassed by training on the combined dataset $\mathcal{G}$, it is also beaten by training on $\mathcal{Z}$, where $|\mathcal{Z}| < |\mathcal{N}|/2$, underscoring its quality.}
\label{tbl:dataset_insights}
\end{table*}

In words, this means that, if the presence of label $j$ indicates that label $i$ would occur with a certain probability for that same instance, then we assume that this probability is also how likely that label $i$ is to be relevant to a data
point that consists of the label features of label $j$.
The right side of \eqref{eq:label-feature-annotation} can be written as
\begin{equation}
    \probability[Y_i=1 \mid Y_j=1] = \probability[Y_i=1, Y_j=1] /
\probability[Y_j=1] \,.
\end{equation}
Thus, we can use the co-occurrence statistics $G_{ij}
\coloneqq \probability[Y_i=1, Y_j=1]$ to calculate the conditionals, and thus
apply a plug-in approach using empirical co-occurrence:
\begin{equation*}
    \probability[Y_i=1 \mid Y_j=1] \approx \frac{\hat{G}_{ij}}{\hat{G}_{jj}} \,, \text{where}, \hat{G}_{ij} \coloneqq \sum_{s=1}^n y_{si} y_{sj}\,.
\end{equation*}
Of course, in the data-scarce XMC regime, the co-occurrence matrix $\boldsymbol{G}$ will be  very noisy. In practice, we empirically find it beneficial to threshold the soft labels at $\delta$, 
so that label features as data-points are annotated by:
\begin{equation}
y_{ij}^{\mathrm{G}} \coloneqq
\begin{cases} 
      \hat{G}_{ij} / \hat{G}_{jj} \text{   if   } & \hat{G}_{ij} / \hat{G}_{jj} > \delta\\
      0 & \text{otherwise}
   \end{cases}
\label{alg:gandalf} \,.
\end{equation}

By approximating the left-hand side of \eqref{eq:label-feature-annotation} using
a parameterized model $\surrogatemodel$, and taking the empirical co-occurrence
as a noise estimate for the right-hand side, we can turn this equation into a
(surrogate) machine-learning task. 
This is the same problem as the original XMC task \eqref{eq:xmc-task}, applied
to a different dataset $\mathcal{Z} = \{(\labelfeature[i], \smash{\labelvec[i]^{\mathrm{G}}} )\}_{i=1}^{\numlabels}$.
That is, we want to optimize
\begin{equation}
    \mathcal{L}_{\mathcal{Z}}[\surrogatemodel, \labeldec] \coloneqq \sum_{i,j=1}^{\numlabels} y_{ij}^{\mathrm{G}}, \langle \surrogatemodel(\labelfeature[j]), \labeldec[i] \rangle) \,.
    \label{eq:surrogate-task}
\end{equation}

In \autoref{tbl:dataset_insights}, we present results for training on this
surrogate task (row \textquote{Training on $\mathcal{Z}$}), when evaluating the
resulting classifier on the original test set. The results are striking, and provide a strong confirmation of the equivalence principle between label features and input texts: Even though this model has \emph{never} seen any
actual training instance, it performs adequate (AmazonTitles) or better
(WikiSeeAlsoTitles) than the original model in terms of precision at $k$. Looking
at PSP, which gives more weight to tail labels, it actually outperforms the original
model, in some cases with a large margin. 

This tells us that we can, in fact, identify the two encoders in equations
\ref{eq:xmc-task} and \ref{eq:surrogate-task}, $\surrogatemodel \equiv
\featureextract$, and train a single model on the combined dataset 
$\mathcal{G} = \mathcal{D} \cup \mathcal{Z}$, as illustrated
in~\autoref{fig:Gandalf}. This combination of data can yield strong improvements
on both regular and propensity scored metrics.

\subsection{Bias-Variance Trade-off}
This improvement cannot be explained by the increased training set size $|\mathcal{G}| = N+L$
alone, as we can show with the following simple experiment: We generate a new dataset
$\mathcal{G}' \sim \mathcal{U}(\mathcal{G}, N)$ by uniformly sampling (without replacement)
from the combined dataset a subset that has the same size as the original training
set $|\mathcal{D}| = N$. \autoref{tbl:dataset_insights} shows that this already 
leads to significant improvements over the original training set.

To explain this phenomenon, we note that this augmented data is qualitatively
slightly different from the original training instances: the empirical
co-occurrence matrix $\hat{\boldsymbol{G}}$ provides \emph{soft} labels
$\labelvec[i]^{\mathrm{G}}$ as training targets.
XMC dataset exhibit high variance~\citep{ProXML,anonymous2024enhancing} because of the long tail labels, whereas the soft labels
of the augmented points provide a much smoother training signal. On the other hand, they are based 
on the approximation of Postulate~\ref{prop:soft-self-annot}, and as such, will introduce some
additional bias into the method, essentially leading to a highly favourable bias-variance trade-off.

In fact, the reduction in variance is so helpful to the training process that
even switching out \eqref{eq:label-feature-annotation} with one-hot labels based
purely on the self-annotation principle ($y_{ij}^{\mathrm{SA}} \coloneqq \indicator[i = j]$ such that
$\mathcal{Z}^1 = \{(\labelfeature[i], \smash{\labelvec[i]^{\mathrm{SA}}} )\}_{i=1}^{\numlabels}$),
thus considerably increasing the bias in the generated data, we still get significant
improvements over just using the original training data (\autoref{tbl:dataset_insights}).

\subsection{Connection to GLaS regularization}

\newcommand{\gdtgt}{\text{GLaS}}

In order to derive a model for $\mathbb{P}[Y_{l'} = 1 \mid X = \labelfeature[l]]$, we can take inspiration from the
\textsc{Glas} regularizer \citep{Guo2019}. This regularizer tries to make the Gram matrix of the label embeddings $\langle \labeldec[i], \labeldec[j] \rangle$
reproduce the co-occurrence statistics of the labels $\mathbf{S}$, 
\begin{equation}
    \mathcal{R}_{\text{GLaS}}[\labeldec] = \numlabels^{-2} \sum_{i=1}^{\numlabels} \sum_{j=1}^{\numlabels} \left(\langle \labeldec[i], \labeldec[j] \rangle - S_{ij} \right)^2.
    \label{eq:glas}
\end{equation}
Here, $\mathbf{S}$ denotes the symmetrized conditional probabilities,
\begin{equation}
\begin{aligned}
    S_{ij} &\coloneqq 0.5 (\mathbb{P}[Y_i=1 \mid Y_{j}=1] + \mathbb{P}[Y_{j}=1 \mid Y_i=1]) \label{glas:similarity}\\
            &\approx 0.5 (\hat{G}_{ij} / \hat{G}_{jj} + \hat{G}_{ij} / \hat{G}_{ii}) \,.
\end{aligned}
\end{equation}

By the self-proximity postulate~\citep{Siamese}, we can assume $\labeldec[l] \approx \featureextract(\labelfeature[l])$.
For a given label feature instance with target soft-label $(\labelfeature[l], y^{\gdtgt}_{ll'}) $, the training will try 
to minimize $\ell(\langle \featureextract(\labelfeature[l]), \labeldec[l']\rangle, y^{\gdtgt}_{ll'})$.
To be consistent with \autoref{eq:glas}, we therefore want to choose $y^{\gdtgt}_{ll'}$ such that 
$S_{ll'} = \argmin \ell(\cdot, y^{\gdtgt}_{ll'})$.
This is fulfilled for $y^{\gdtgt}_{ll'} = \sigma(S_{ll'})$ for $\ell$ being the binary cross-entropy, where
$\sigma$ denotes the logistic function.\looseness=-1

While the soft targets generated this way slightly differ from the ones of \eqref{alg:gandalf},
as already observed, the bias introduced by mildly incorrect training targets is
offset by far by the variance reduction, and we find that this version performs
only slightly worse than \emph{Gandalf} (\autoref{visualisations}).

\newpage
\section{Experiments}

\paragraph{\textbf{Benchmarks, Baseline and Metrics}} We benchmark our experiments on 4 standard public datasets, the details of which are mentioned in \autoref{tab:datasets}.
To test the generality and effectiveness of our proposed \textit{Gandalf}, we apply the algorithm across a variety of state-of-the-art short-text extreme classifiers. These consist of (i) base frugal models -\textsc{Astec} \cite{Astec} and \textsc{InceptionXML} \cite{kharbanda2021embedding} - which do not, by default, leverage label text information, (ii) \textsc{Decaf} \cite{Decaf}, \textsc{Eclare} \cite{Eclare} and \textsc{InceptionXML-LF} which equip the base models with additional encoders to make use of label text and label correlation information and, (iii) \textsc{Ngame + Renee} - consisting of \textsc{Renee} \cite{renee}, which makes CUDA optimizations to train BCE loss over a classifier for $L$ labels without a shortlist. The transformer encoder is initialized with pre-trained \textsc{Ngame} (M1, dual encoder) \cite{NGAME}. 
We measure the performance using standard metrics P@k, its propensity-scored variant, PSP@k \citep{jain2016extreme,qaraei2021convex}, and coverage@k \cite{schultheis2022missing, schultheis2024generalized}.
\begin{table*}[!h]
\begin{adjustbox}{width=\textwidth,center}
\begin{tabular}{c|ccc|ccc|ccc|ccc}
\toprule
\textbf{Method}     & \textbf{P@1}   & \textbf{P@3}   & \textbf{P@5}   & \textbf{PSP@1}   & \textbf{PSP@3}   & \textbf{PSP@5} & \textbf{P@1}   & \textbf{P@3}   & \textbf{P@5}   & \textbf{PSP@1}   & \textbf{PSP@3}   & \textbf{PSP@5} \\
\specialrule{0.70pt}{0.4ex}{0.65ex}
& \multicolumn{6}{c|}{\textbf{LF-AmazonTitles-131K}} & \multicolumn{6}{c}{\textbf{LF-AmazonTitles-1.3M}} \\
\specialrule{0.70pt}{0.4ex}{0.65ex}
\textsc{SiameseXML} & 41.42 & \textbf{30.19} & 21.21 & 35.80  & 40.96 & 46.19 & 49.02 & 42.72 & 38.52 & 27.12 & 30.43 & 32.52\\ 
\midrule
\textsc{Astec} & 37.12 & 25.20 & 18.24 & 29.22 & 34.64 & 39.49 & 48.82 & 42.62 & 38.44 & 21.47 & 25.41 & 27.86\\
+ \textit{Gandalf} & 43.95 & 29.66 & 21.39 & 37.40 & 43.03 & 48.31 & 53.02 & 46.13 & 41.37 & 27.32 & 31.20 & 33.34\\
\midrule
\textsc{Decaf} & 38.40  & 25.84 & 18.65 & 30.85 & 36.44 & 41.42 & 50.67 & 44.49 & 40.35 & 22.07 & 26.54 & 29.30\\
+ \textit{Gandalf} & 42.43 & 28.96 & 20.90  & 35.22 & 42.12 & 47.61 & 53.02 & 46.65 & 42.25 & 25.47 & 30.14 & 32.83\\
\midrule
\textsc{Eclare} & 40.46 & 27.54 & 19.63 & 33.18 & 39.55 & 44.10 & 50.14 & 44.09 & 40.00 & 23.43 & 27.90 & 30.56\\
+ \textit{Gandalf} & 42.51  & 28.89  & 20.81 & 35.72 & 42.19 & 47.46 & 53.87 & 47.45 & 43.00 & 28.86 & 32.90 & 35.20\\
\midrule
\textsc{InceptionXML}& 36.79 &  24.94 &  17.95 & 28.50 &  34.15 & 38.79 & 48.21 & 42.47 & 38.59 & 20.72 & 24.94 & 27.52\\
+ \textit{Gandalf} & 44.67 & 30.00 & 21.50 & 37.98 & 43.83 & 48.93 & 50.80 & 44.54 & 40.25 & 25.49 &  29.42 & 31.59\\
\midrule
\textsc{InceptionXML-LF} & 40.74 & 27.24 & 19.57 & 34.52 & 39.40 & 44.13 & 49.01 & 42.97  & 39.46 & 24.56  & 28.37 & 31.67\\
+ \textit{Gandalf} & 43.84 & 29.59 & 21.30 & 38.22 & 43.90 & 49.03 & 52.91 & 47.23 & 42.84 & \textbf{30.02} & 33.18 & 35.56\\
\midrule
\textsc{Ngame + Renee} & \textbf{46.05} & \textbf{30.81} & \textbf{22.04} & 38.47 & 44.87 & 50.33 & 56.04 & 49.91 & 45.32 & 28.54 & \textbf{33.38} & \textbf{36.14}\\
+ \textit{Gandalf} & 45.86 & 30.53 & 21.79 & \textbf{40.49}& \textbf{45.83} & \textbf{50.96} & \textbf{56.88} & \textbf{50.24} & \textbf{45.47} & 26.56 & 31.69 & 34.60\\

\specialrule{0.70pt}{0.4ex}{0.65ex}
    & \multicolumn{6}{c|}{\textbf{LF-WikiSeeAlsoTitles-320K}} & \multicolumn{6}{c}{\textbf{LF-WikiTitles-500K}} \\
\specialrule{0.70pt}{0.4ex}{0.65ex}
\textsc{SiameseXML}  & 31.97 & 21.43 & 16.24 & \textbf{26.82} & 28.42 & 30.36 & 42.08 & 22.80 & 16.01 & 23.53 & 21.64 & 21.41\\
\midrule
\textsc{Astec} & 22.72 & 15.12 & 11.43 & 13.69 & 15.81 & 17.50 & 44.40 & 24.69 & 17.49 & 18.31 & 18.25 & 18.56\\
+ \textit{Gandalf} & 31.10 & 21.54 & 16.53 & 23.60 & 26.48 & 28.80 & 45.24 & 25.45 & 18.57 & 21.72 & 20.99 & 21.16 \\
\midrule
\textsc{Decaf} & 25.14 & 16.90  & 12.86 & 16.73 & 18.99 & 21.01 & 44.21 & 24.64 & 17.36 & 19.29 & 19.82 & 19.96\\
+ \textit{Gandalf}  & 31.10 & 21.60 & 16.31 & 24.83 & 27.18 & 29.29 & 45.27 & 25.09 & 17.67 & 22.51 & 21.63 & 21.43\\
\midrule
\textsc{Eclare} & 29.35 & 19.83 & 15.05  & 22.01 & 24.23 & 26.27 & 44.36 & 24.29 & 16.91 & 21.58 & 20.39 & 19.84\\
+ \textit{Gandalf} & 31.33 & 21.40 & 16.31 & 24.83 & 27.18 & 29.29 & 45.12 & 24.45 & 17.05 & 24.22 & 21.41 & 20.55\\
\midrule
\textsc{InceptionXML} & 23.10 & 15.54 & 11.52 & 14.15 & 16.71 & 17.39 & 44.61 & 24.79 & 19.52 & 18.65 & 18.70 & 18.94 \\
+ \textit{Gandalf} & 32.54 & 22.15 & 16.86 & 25.27 &  27.76 & 30.03 & 45.93  & 25.81 & \textbf{20.36} & 21.89  & 21.54 & 22.56\\
\midrule
\textsc{InceptionXML-LF} & 28.99 & 19.53 & 14.79  & 21.45 & 23.65 & 25.65 & 44.89 & 25.71 & 18.23 & 23.88 & 22.58 & 22.50\\
+ \textit{Gandalf} & 33.12 & 22.70  & 17.29 & 26.68 & \textbf{29.03} &  \textbf{31.27} & \textbf{47.13} & \textbf{26.87} & 19.03 & \textbf{24.12} & \textbf{23.92} & \textbf{23.82}\\
\midrule
\textsc{Ngame + Renee} & 30.79 & 20.65 & 15.57 & 20.81 & 24.46 & 27.05 & - & - & - & - & - & -\\
+ \textit{Gandalf} & \textbf{33.92} & \textbf{23.11} & \textbf{17.58} & 24.15 & 26.23 & 30.89 & - & - & - & - & - & -\\
\bottomrule
\end{tabular}
\end{adjustbox}
\caption{Results showing the effectiveness of \textit{Gandalf} on state-of-the-art extreme classifiers. The best results are in \textbf{bold}.}
\label{tbl:main_results_dual}
\vspace{-2em}
\end{table*}

\subsection{Main Results}

\paragraph{\textbf{Gandalf vs Architectural Additions (\textit{LTE}, \textit{GALE})}} The first formal attempt to externally imbue the model with label information was made with \textsc{Decaf}, which essentially equips the base model \textsc{Astec} with another base encoder (\textit{LTE}) to learn label text ($\labelfeature[l]$) embeddings along with the classifier. 
The second attempt, in the form of \textsc{Eclare}, builds upon \textsc{Decaf} by adding another base encoder (\textit{GALE}) to process and \textit{externally} capture label correlation information. 
To make our claim more general, we also evaluate on \textsc{InceptionXML-LF}, which consist of the same extensions on a more recent base model \textsc{InceptionXML} \cite{kharbanda2021embedding} with \textit{LTE} and \textit{GALE} components (just as \textsc{Eclare} adds \textit{LTE} and \textit{GALE} to \textsc{Astec}). 
While such architectural modifications help capture higher order query-label relations and increase empirical performance, they also increase both training time and the peak GPU memory required during training by $\sim 3\times$.

\begin{figure}[t]
    \centering
  \includegraphics[trim={0 0 0 1.15cm}, clip, width=\columnwidth]{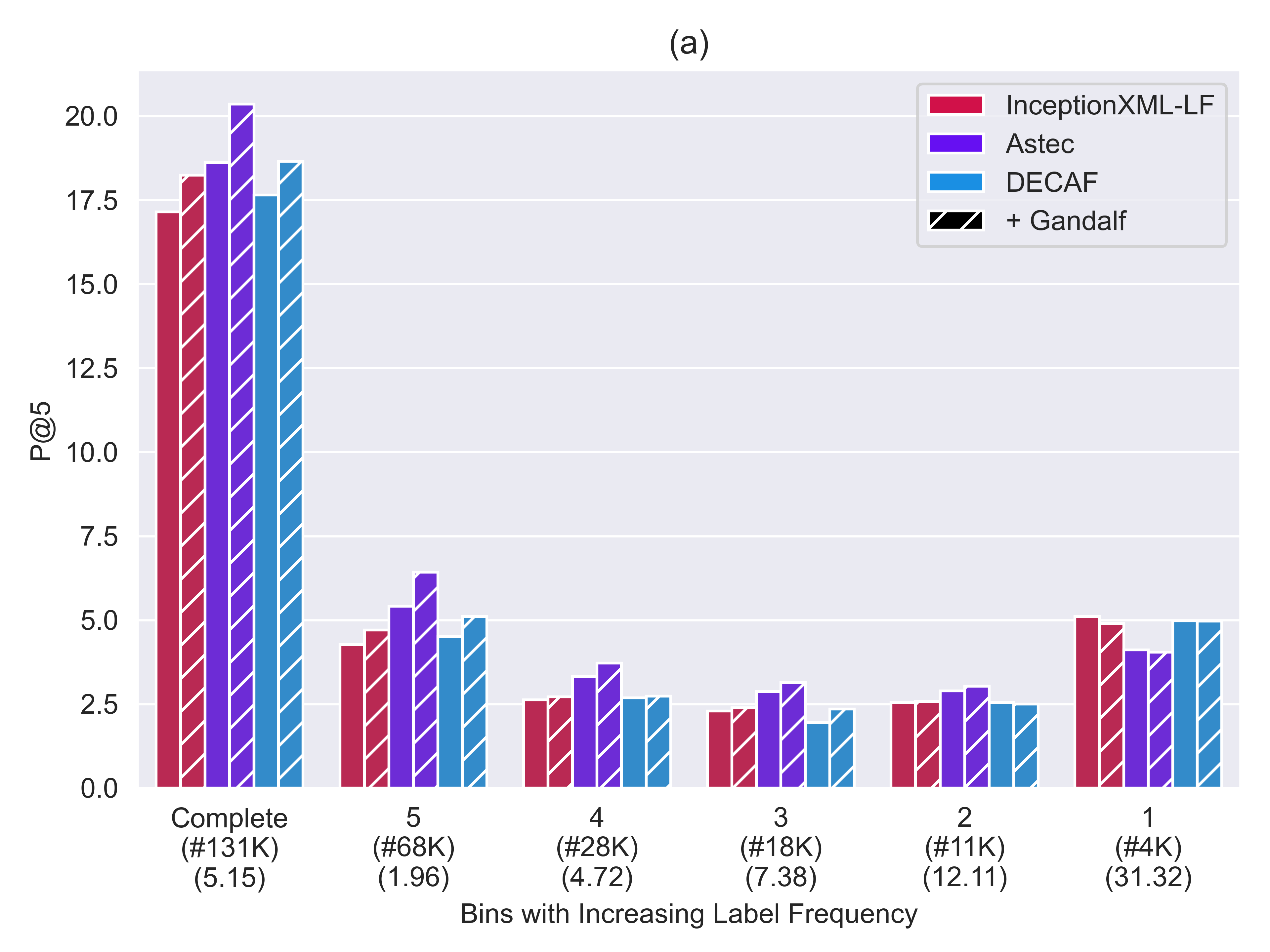}
    \caption{\textsc{Gandalf} demonstrating improvements on the P@5 metric across various methods, separated into tail, torso and head lables. On the x axis, the middle row indicates the number of labels in the bin, and the lowest row denotes the average number of positives per label in that bin. Improvements in earlier bins (5 - 3) denote gains in tail label performance.}
    \label{fig:teaser}
    \vspace{-1em}
\end{figure}

As \textit{Gandalf} is a data-centric approach, the memory overhead is eliminated by default. Further, we find that (i) \textsc{Decaf} and \textsc{Eclare} still benefit from using \textit{Gandalf} augmented data implying architectural modifications are complementary to \textit{Gandalf}. However, (ii) simply using \textit{Gandalf} augmented data enables base models \textsc{Astec} and \textsc{InceptionXML} outperform themselves by up to 30\% and perform nearly at par with their more architecturally equipped counter parts \textsc{Eclare} and \textsc{InceptionXML-LF}. 
While we posit that \textit{Gandalf} and \textit{GALE} learn complementary data relations, both our quantitative (\autoref{tbl:main_results_dual}) and qualitative (\autoref{tbl:main_comparisons}, \autoref{fig:teaser}) results show that \textit{Gandalf} is more effective and efficient at capturing these relations (specifically, label correlations) compared to the latter. 

\paragraph{\textbf{Gandalf vs Siamese Learning}} Consequently, the third attempt made at capturing label correlations via \textsc{SiameseXML}, which essentially replaces the surrogate training task in \textsc{Astec} with a two-tower siamese learning framework. As argued in \autoref{sec:preliminaries}, the condition $\labeldec[l] = \featureextract(\labelfeature[l])$ turns out to be too strong, and consequently training of \textsc{SiameseXML} and \textsc{NGAME} is done in two
stages. Initially, a contrastive loss needs to be minimized, followed by
fine-tuning with a classification objective which allows for some fine-tuning corrections $\boldsymbol \eta_l$, yielding $\labeldec[l] = \featureextract(\labelfeature[l]) + \boldsymbol \eta_l$. 
On the other hand, \textit{Gandalf} simply extends training data to learn from a-priori label co-occurrence data in a supervised manner. 
Notably (from \autoref{tbl:main_results_dual}), \textsc{Astec} + \textit{Gandalf} outperforms \textsc{SiameseXML} by 5-10\% on Amazon datasets, while performing at par on Wikipedia datasets. 

\paragraph{\textbf{Applying Gandalf to Two-tower approaches}} 
Although we propose \textsc{Gandalf} as a method suitable for training classifiers, it can also be used leveraged alongside two-tower approaches, like \textsc{Ngame}. This is done by first extending the dual encoder with a scalable classifier with \textsc{Renee}, which simply trains OvA classifiers on top of the base model. Using \textsc{Gandalf} augmented data during this extension leads to significant improvements, more prominently on the LF-WikiSeeAlsoTitles-320K and LF-AmazonTitles-131K datasets.

\paragraph{\textbf{Improvements on tail labels}} We perform a quantile analysis across 2 datasets -- LF-AmazonTitles-131K (\autoref{fig:teaser}) and the LF-WikiSeeAlsoTitles-320K (\autoref{fig: freq_dist}, \autoref{visualisations}) with \textsc{InceptionXML} -- where we examine performance (contribution to P@5 metric) over 5 equi-voluminous bins based on increasing order of mean label frequency in the training dataset. Consequently, performance on head labels can be captured by the bin \#1 and that of tail labels by bin \#5. We note that introducing the additional training data with \textit{Gandalf} consistently improves the performance across all label frequencies, with more profound gains on bins with more tail labels. This is further verified by significant performance boosts, with base models showing upto 11\% improvements in the PSP@k metrics in \autoref{tbl:main_results_dual}. 

\paragraph{\textbf{Beyond model performances}} We can also extract dataset specific insights with \textsc{Gandalf} from \autoref{tbl:main_results_dual}. Significant improvements on top of the base algorithm are particularly observed on LF-AmazonTitles-131K and LF-WikiSeeAlsoTitles-320K. In contrast, improvements on LF-WikiTitles-500K remain relatively mild. We attribute this to the density of the datasets. Specifically, while the former datasets consist of $\sim$5 training instances per label, the latter consists of $\sim$17. We posit a higher query-label density enables algorithms to inherently learn sufficient label-label correlations from existing data. 
However, we further see that using \textit{Gandalf} is effective for LF-AmazonTitles-1.3M, the largest public benchmark for XMC with label features. Here, even though average training instances per label is $\sim$38, the average number of labels per instance is $\sim$22, as compared to maximum of $\sim$4 on other datasets.

\paragraph{\textbf{Qualitative Results}}
We further analyse qualitative examples via the top 5 predictions obtained by training the base encoders with and without \textit{Gandalf} augmented data points in \autoref{tbl:main_comparisons}, with more examples in \autoref{visualisations}. Notably, we can observe that queries with even a single keyword (\textit{Oat}), which have no correct predictions without \textsc{Gandalf}, result in 100\% correct predictions with it. Furthermore, even the quality of incorrect predictions improves \footnote{These labels are not annotated, but are most likely \textit{missed true positives} \cite{jain2016extreme}}. For example, in case of ``Lunar Orbiter program'', the only incorrect \textit{Gandalf} predictions are ``Lunar Orbiter 3'', ``Lunar Orbiter 5'' and ``Pioneer program'' (US lunar and planetary space programs).

Additionally, we show semantic similarity between the annotated labels with $\mathcal{G}$, and the original label in \autoref{fig:LCG_visual} in \autoref{visualisations}.
\vspace{-2em}

\begin{table*}[h!]
\centering
\begin{adjustbox}{width=\textwidth, center}
\begin{tabular}{|p{3cm}|p{2cm}|p{9.4cm}p{7cm}|}
\toprule
\multicolumn{1}{|c|}{\textbf{Method}}     & \multicolumn{1}{c|}{\textbf{Datapoint}}  & \multicolumn{1}{c}{\textbf{Baseline Predictions}} & \multicolumn{1}{c|}{\textbf{\textit{Gandalf} Predictions}}  \\
\specialrule{0.70pt}{0.4ex}{0.65ex}
\multicolumn{1}{|c|}{\multirow{2}{*}{\textsc{InceptionXML-LF}}} & \multicolumn{1}{c|}{\multirow{2}{*}{}} & \textcolor{lightgray}{Pontryagin duality}, \textcolor{lightgray}{Topological order}, \textcolor{lightgray}{Topological quantum field theory}, \textcolor{lightgray}{Topological quantum number}, \textcolor{lightgray}{Quantum topology} & Compact group, \textcolor{lightgray}{Haar measure}, Lie group, Algebraic group, Topological ring \\
\cmidrule{1-1}
\cmidrule{3-4}
\multicolumn{1}{|c|}{\multirow{2}{*}{\textsc{Decaf}}} &  \multicolumn{1}{c|}{\multirow{2}{*}{Topological group}} & \textcolor{lightgray}{Topological quantum computer}, \textcolor{lightgray}{Topological order}, \textcolor{lightgray}{Topological quantum field theory}, \textcolor{lightgray}{Topological quantum number}, \textcolor{lightgray}{Quantum topology} & Compact group, \textcolor{lightgray}{Haar measure}, Lie group, Algebraic group, Topological ring \\
\cmidrule{1-1}
\cmidrule{3-4}
\multicolumn{1}{|c|}{\multirow{2}{*}{\textsc{Eclare}}} & \multicolumn{1}{c|}{\multirow{2}{*}{}} & \textcolor{lightgray}{Topological quantum computer}, \textcolor{lightgray}{Topological order}, \textcolor{lightgray}{Topological quantum field theory}, \textcolor{lightgray}{Topological quantum number}, \textcolor{lightgray}{Quantum topology} & Compact group, \textcolor{lightgray}{Topological order}, Lie group, Algebraic group, Topological ring \\
\midrule
\midrule
\multicolumn{1}{|c|}{\multirow{2}{*}{\textsc{InceptionXML-LF}}} & \multicolumn{1}{c|}{\multirow{2}{*}{}} & \textcolor{lightgray}{List of lighthouses in Scotland}, \textcolor{lightgray}{List of Northern Lighthouse Board lighthouses}, Oatcake, \textcolor{lightgray}{Communes of the Finistere department}, {Oat milk} & Oatcake, Oatmeal, Oat milk, Porridge, Rolled oats \\
\cmidrule{1-1}
\cmidrule{3-4}
\multicolumn{1}{|c|}{\multirow{2}{*}{\textsc{Decaf}}} & \multicolumn{1}{c|}{\multirow{2}{*}{Oat}} & Oatcake, Oatmeal, \textcolor{lightgray}{Design for All (in ICT)}, \textcolor{lightgray}{Oatley Point Reserve}, \textcolor{lightgray}{Oatley Pleasure Grounds} & Oatcake, Oatmeal, Oat milk, Porridge, Rolled oats \\
\cmidrule{1-1}
\cmidrule{3-4}
\multicolumn{1}{|c|}{\multirow{2}{*}{\textsc{Eclare}}} &  \multicolumn{1}{c|}{\multirow{2}{*}{}} & Oatmeal, Oat milk, \textcolor{lightgray}{Parks in Sydney}, \textcolor{lightgray}{Oatley Point Reserve}, \textcolor{lightgray}{Oatley Pleasure Grounds} & Oatcake, Porridge, Rolled oats, \textcolor{lightgray}{Oatley Point Reserve}, \textcolor{lightgray}{Oatley Pleasure Grounds} \\
\midrule
\midrule
\multicolumn{1}{|c|}{\multirow{2}{*}{\textsc{InceptionXML-LF}}} & \multicolumn{1}{c|}{\multirow{2}{*}{}} & Lunar Orbiter Image Recovery Project, \textcolor{lightgray}{Lunar Orbiter 3}, \textcolor{lightgray}{Lunar Orbiter 5}, \textcolor{lightgray}{Chinese Lunar Exploration Program}, \textcolor{lightgray}{List of future lunar missions} &  Surveyor program, Luna programme, Lunar Orbiter Image Recovery Project, \textcolor{lightgray}{Lunar Orbiter 3}, \textcolor{lightgray}{Lunar Orbiter 5} \\
\cmidrule{1-1}
\cmidrule{3-4}
\multicolumn{1}{|c|}{\multirow{2}{*}{\textsc{Decaf}}} & \multicolumn{1}{c|}{\multirow{2}{*}{Lunar Orbiter program}}  & Exploration of the Moon, \textcolor{lightgray}{List of man-made objects on the Moon}, Lunar Orbiter Image Recovery Project, \textcolor{lightgray}{Lunar Orbiter 3}, \textcolor{lightgray}{Lunar Orbiter 5} & Exploration of the Moon, Apollo program, Surveyor program, Luna programme, Lunar Orbiter program \\
\cmidrule{1-1}
\cmidrule{3-4}
\multicolumn{1}{|c|}{\multirow{2}{*}{\textsc{Eclare}}} & \multicolumn{1}{c|}{\multirow{2}{*}{}} & Exploration of the Moon, Lunar Orbiter program, Lunar Orbiter Image Recovery Project, \textcolor{lightgray}{Lunar Orbiter 3}, \textcolor{lightgray}{Lunar Orbiter 5} & Exploration of the Moon, \textcolor{lightgray}{Pioneer program}, Surveyor program, Luna programme, Lunar Orbiter program \\
\midrule
\midrule
\multicolumn{1}{|c|}{\multirow{3}{*}{\textsc{InceptionXML-LF}}} & \multicolumn{1}{c|}{\multirow{3}{*}{}} &  \textcolor{lightgray}{Colorado metropolitan areas}, \textcolor{lightgray}{Front Range Urban Corridor}, Outline of Colorado, Index of Colorado-related articles, \textcolor{lightgray}{State of Colorado} & \textcolor{lightgray}{Colorado metropolitan areas}, Outline of Colorado, Index of Colorado-related articles, Colorado cities and towns, Colorado counties \\
\cmidrule{1-1}
\cmidrule{3-4}
\multicolumn{1}{|c|}{\multirow{3}{*}{\textsc{Decaf}}} & \multicolumn{1}{c|}{\multirow{3}{*}{Grand Lake, Colorado}}  & \textcolor{lightgray}{Colorado metropolitan areas}, \textcolor{lightgray}{Front Range Urban Corridor}, \textcolor{lightgray}{State of Colorado}, \textcolor{lightgray}{Colorado municipalities}, \textcolor{lightgray}{National Register of Historic Places listings in Grand County,  Colorado} & Outline of Colorado, \textcolor{lightgray}{State of Colorado}, Colorado cities and towns, \textcolor{lightgray}{Colorado municipalities}, Colorado counties \\
\cmidrule{1-1}
\cmidrule{3-4}
\multicolumn{1}{|c|}{\multirow{3}{*}{\textsc{Eclare}}} & \multicolumn{1}{c|}{\multirow{3}{*}{}} & \textcolor{lightgray}{State of Colorado}, Colorado cities and towns, Colorado counties, \textcolor{lightgray}{National Register of Historic Places listings in Grand County,  Colorado}, \textcolor{lightgray}{Grand County,  Colorado} & Outline of Colorado, Index of Colorado-related articles, \textcolor{lightgray}{State of Colorado}, Colorado cities and towns, Colorado counties \\
\specialrule{0.70pt}{0.4ex}{0.65ex}
\end{tabular}
\end{adjustbox}
\caption{Qualitative predictions from the LF-WikiSeeAlsoTitles-320K dataset. \textcolor{lightgray}{Labels} indicate mispredictions.}
\label{tbl:main_comparisons}
\vspace{-2em}
\end{table*}
\subsection{Ablation \& Computational Analysis}
\textit{Gandalf}, is a data-centric approach that does not increase the computational cost during inference. While the inclusion of label features - which can often run in the order of millions - as additional data points might seem to increase the computational cost during training, through a series of observations, we show that this is in fact not the case. On the contrary, \textit{Gandalf} can help in reducing the memory footprint while training, enabling researchers to use smaller GPUs, and reallocating their compute budget towards longer training schedules. Secondly, we also study the effect of subsampling the labels used for \textit{Gandalf} to demonstrate how learning even some of the label-label correlations is beneficial for XMC models. This observation is particularly useful when inclusion of all label-features as data points becomes intractable due to its scale.

\begin{figure*}
    \centering
    \subfigure[]{\includegraphics[trim={0 0 0 1.15cm}, clip, width=0.245\textwidth]{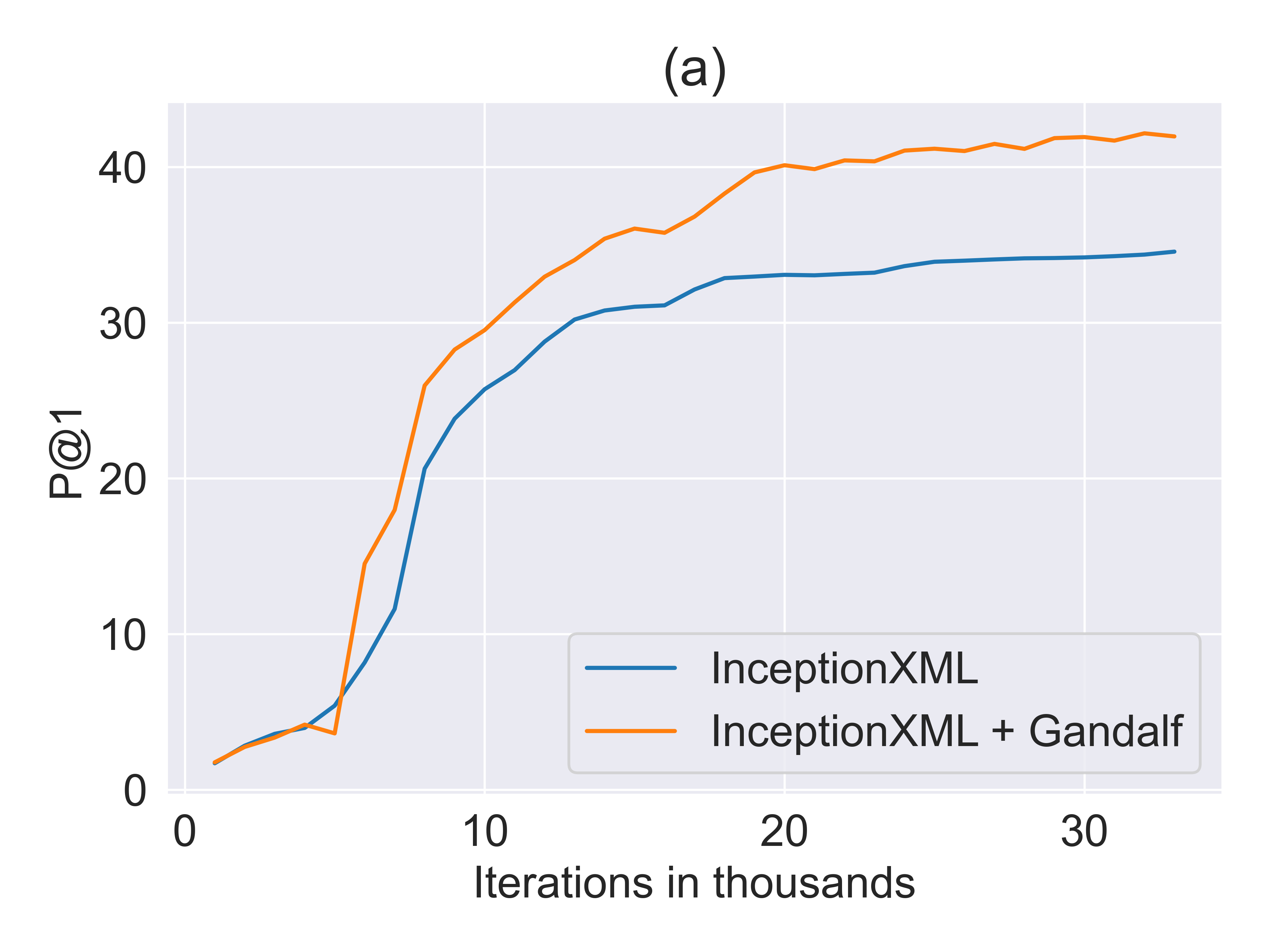}} 
    \subfigure[]{\includegraphics[trim={0 0 0 1.15cm}, clip, width=0.245\textwidth]{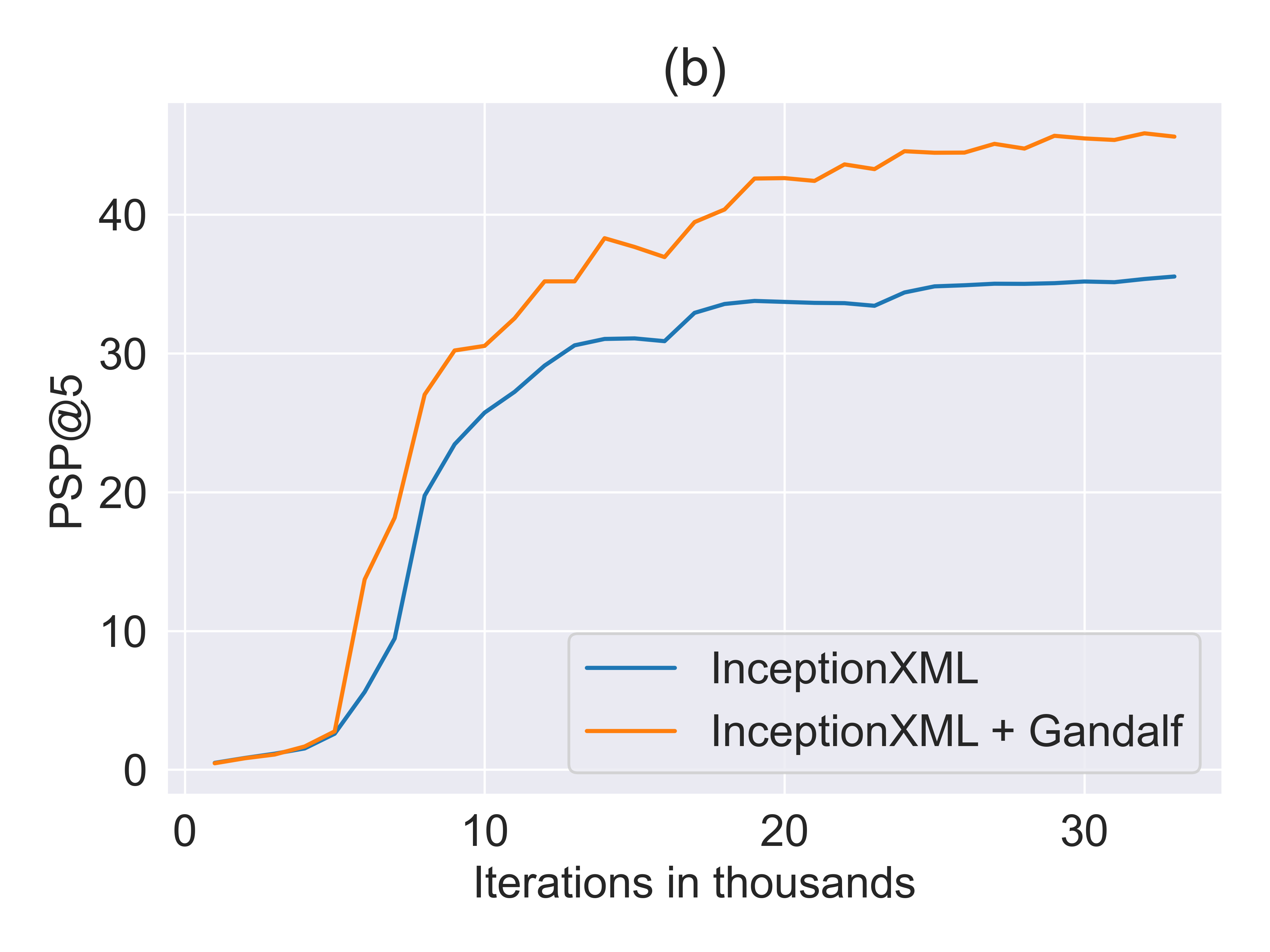}} 
    \subfigure[]{\includegraphics[trim={0 0 0 1.15cm}, clip, width=0.245\textwidth]{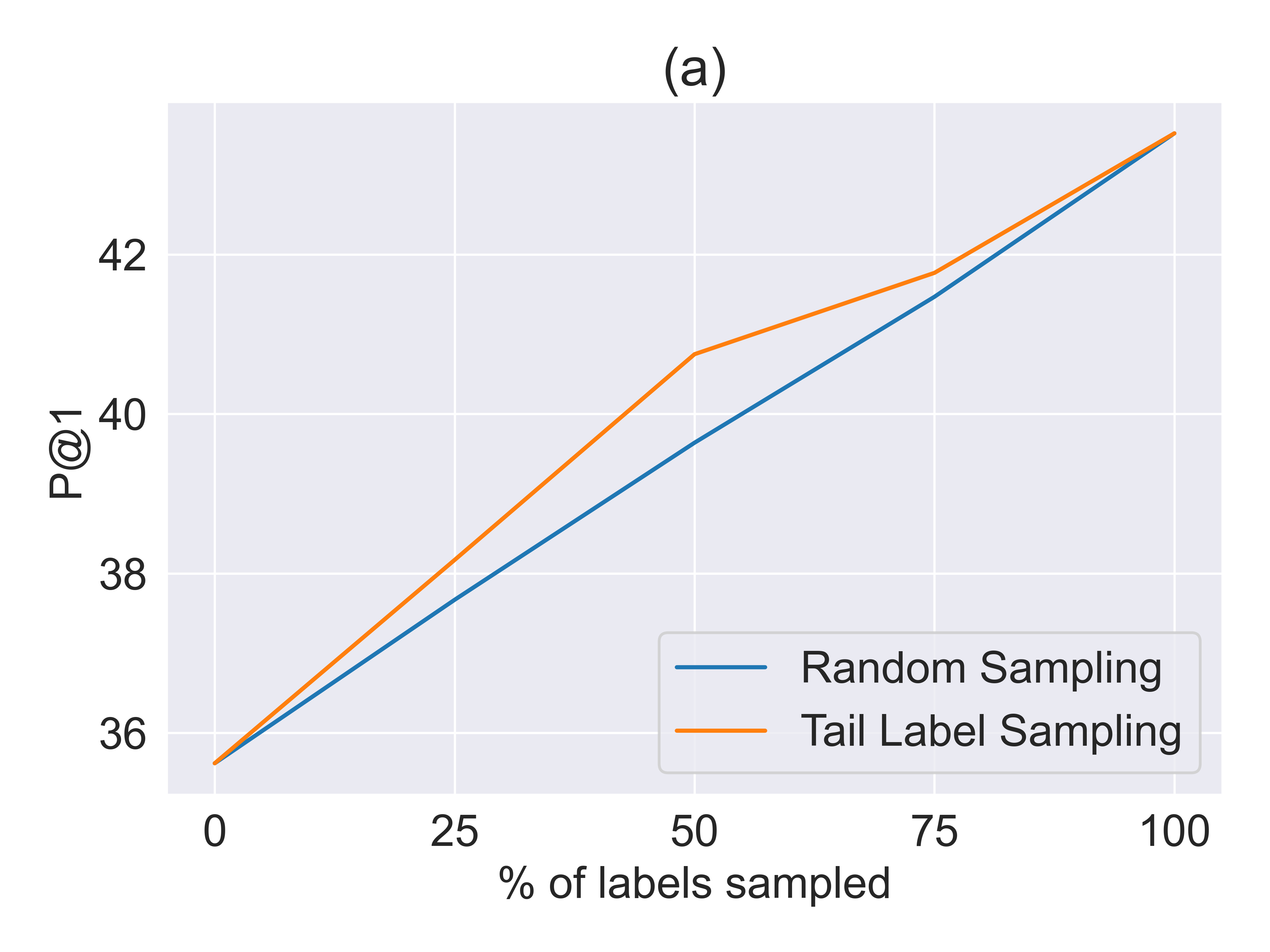}}
    \subfigure[]{\includegraphics[trim={0 0 0 1.15cm}, clip, width=0.245\textwidth]{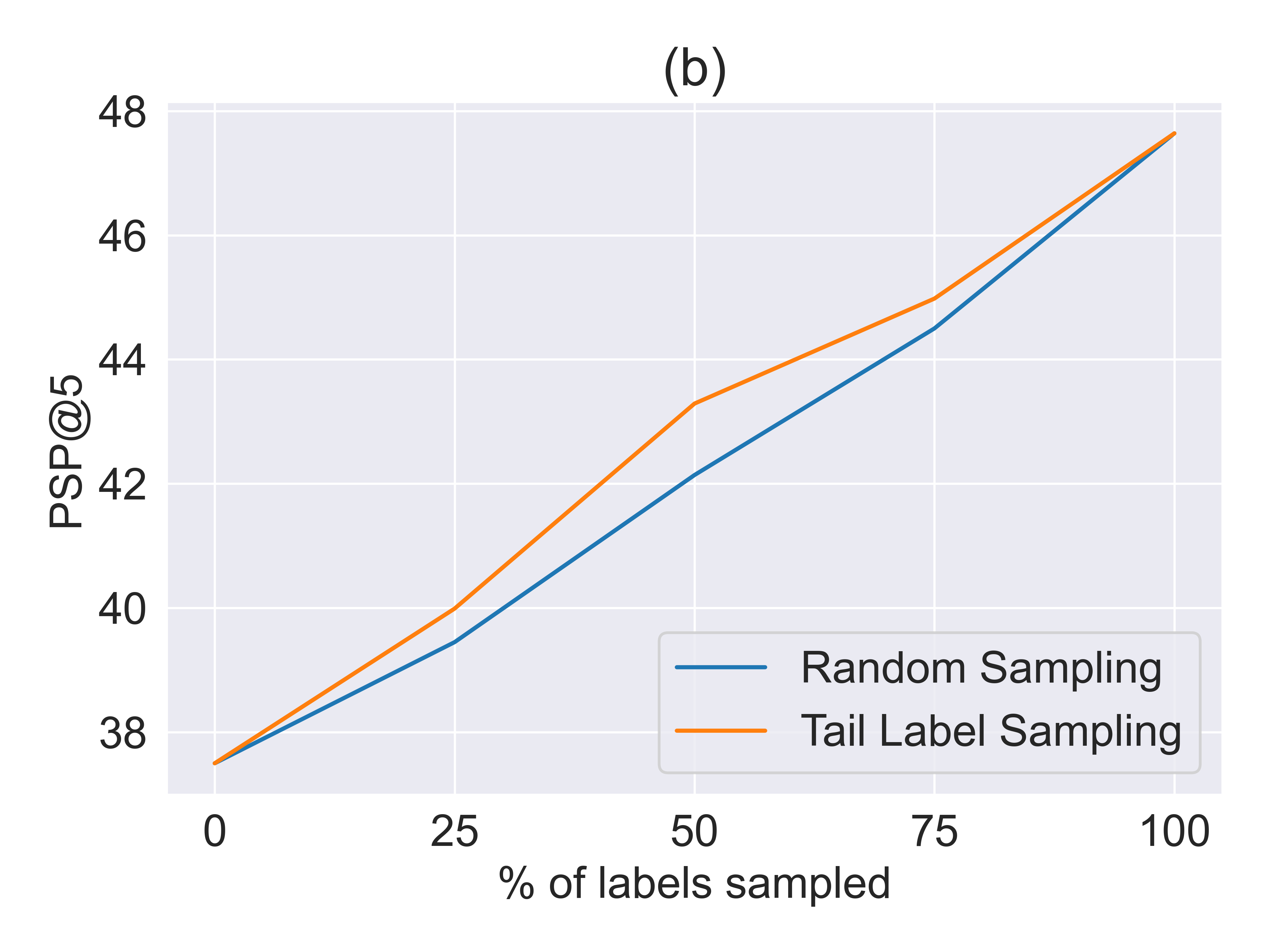}}
    \vspace{-1em}
    \caption{The (a) P@1 and (b) PSP@5 metric plotted against iterations for InceptionXML with and without \textit{Gandalf}. The effect of subsampling labels for \textit{Gandalf} on the (c) P@1 and (d) PSP@5 metric. Both results are on the LF-AmazonTitles-131K dataset.}
    \label{fig: iterations}
\end{figure*}

\begin{table*}[th]
\begin{adjustbox}{width=\textwidth,center}
\begin{tabular}{c|ccc|ccc|ccc|ccc}
\toprule
\textbf{Method} & \textbf{P@1}   & \textbf{P@3}   & \textbf{P@5}   & \textbf{PSP@1} & \textbf{PSP@3} & \textbf{PSP@5} & \textbf{P@1}   & \textbf{P@3}   & \textbf{P@5}   & \textbf{PSP@1} & \textbf{PSP@3} & \textbf{PSP@5} \\
\specialrule{0.70pt}{0.4ex}{0.65ex}
    & \multicolumn{6}{c|}{\textbf{LF-AmazonTitles-131K}} & \multicolumn{6}{c}{\textbf{LF-WikiSeeAlsoTitles-320K}} \\
\specialrule{0.70pt}{0.4ex}{0.65ex}
InceptionXML & 35.62 & 24.13 & 17.35 & 27.53 & 33.06 & 37.50 & 21.53 & 14.19 & 10.66 & 13.06 & 14.87 & 16.33 \\
+ \textit{Gandalf $\mathcal{G}$}    & \textbf{43.71} & \textbf{29.30} & \textbf{21.14} & \textbf{37.25} & \textbf{43.01} & \textbf{47.89}   & \textbf{31.42} & \textbf{21.54} & \textbf{16.37} & \textbf{24.78} & \textbf{27.36} & \textbf{28.98}\\
+ \textit{Gandalf ($\mathcal{G}$ + Random Walk \cite{Eclare})} & 43.52 & 29.23 & 20.92 & 36.96 & 42.71 & 47.64   & 31.31 & 21.38 & 16.22 & 24.31 & 26.79 & 28.83\\
\midrule
\textsc{InceptionXML-LF} & 40.74 & 27.24 & 19.57 & 34.52 & 39.40 & 44.13 & 49.01 & 42.97  & 39.46 & 24.56  & 28.37 & 31.67\\
+ \textit{Gandalf} ($\delta$ = 0.0) & 41.71 & 28.03 & 20.14 & 36.94 & 41.93 & 46.64 & 31.40 & 21.56 & 16.53 & 26.01 & 27.89 & 29.99\\
+ \textit{Gandalf} ($\delta$ = 0.1) & \textbf{42.09} & \textbf{28.38} & \textbf{20.45} & \textbf{37.09} & \textbf{42.19} & \textbf{47.04} & \textbf{32.20} & \textbf{21.86} & \textbf{16.60} & \textbf{26.06} & \textbf{28.01} & \textbf{30.03}\\
+ \textit{Gandalf} ($\delta$ = 0.2) & 41.73 & 28.10 & 20.18 & 37.01 & 41.99 & 46.67 & 31.29 & 21.35 & 16.28 & 25.68 & 27.59 & 29.65\\
\bottomrule
\end{tabular}
\end{adjustbox}
\vspace{0.1in}
\caption{Results demonstrating the effectiveness of \textit{Gandalf} using both, a statistical co-occurrence matrix ($\mathcal{G}$) and it's modified version using a random walk as in \cite{Eclare}. The table also shows the method's sensitivity to $\delta$, as defined in Equation \ref{alg:gandalf}.}
\label{tbl:ablation_smooth_labels}
\vspace{-0.2in}
\end{table*}

\paragraph{\textbf{Computational Costs during Training}} For the LF-Amazon-
Titles-131K dataset, we plot the P@1 and the PSP@5 metric against iterations for InceptionXML, trained with and without Gandalf in \autoref{fig: iterations}. As can be seen, using \textit{Gandalf} gives better performance, even on tail labels, right from the beginning. Moreover, where the performance of InceptionXML saturates, the performance of \textit{Gandalf} continues to scale with increasing compute. Therefore, given a fixed computational budget, a model trained with \textit{Gandalf} will outperform one trained without it. This can also be seen in \autoref{tbl:dataset_insights} where training on $\mathcal{U}(\mathcal{G}, N)$, i.e., under the exact same computational budget as training on the the original dataset gives performance improvements. In the same table, we can also observe improvements when training on \textit{less than half} the original compute with $\mathcal{Z}$ \footnote{These notations have been defined in \autoref{sec:gandalf}}.
These observations firmly place \textit{Gandalf} as a compute-efficient method of leveraging label-features in XMC models.\footnote{Note that the creation of the LxL label correlation graph takes less than two minutes, even for the large LF-AmazonTitles-1.3M dataset. This is only done once before training and has a negligible effect on the computational cost.}

\paragraph{\textbf{Effect of Subsampling Labels}} We demonstrate the effect of subsampling labels used for \textit{Gandalf} under two schemes, (a) Randomly sampling an expected percentage subset of labels and (b) randomly sampling this subset from equi-voluminous bins of increasing label frequency, i.e., prioritising tail labels for lower percentages. These results are shown for the P@1 and PSP@5 metric on the LF-AmazonTitles-131K dataset in \autoref{fig: iterations}.

Both the metrics grow linearly as the percentage sampled labels are increased in steps of 25\%. This goes ahead to show the lack of label-label correlations being captured in existing methods, and how learning even on a subset can be useful. Further, prioritising tail-labels consistently outperforms the random sampling baseline, underscoring the data-scarcity issue in XMC.

\paragraph{\textbf{Choice of label co-occurrence graph $\mathcal{G}$}} While with \textsc{Gandalf}, we leverage a statistical measure for $\mathcal{G}$, we can also estimate it with random walks \cite{Eclare} (used for \textit{GALE}). We find that our method is not significantly affected by this choice, with the co-occurrence graph giving slightly enhanced performance(\autoref{tbl:ablation_smooth_labels}). We hypothesise this happens due to the noise introduced via random walks. While both variants aim to model similar information, their differing usage determines their overall effectiveness. In particular, leveraging it for \textsc{Gandalf} helps learn sufficient information on top of \textit{GALE}. 

\paragraph{\textbf{Sensitivity to $\delta$}}
We examine \textit{Gandalf}'s sensitivity to $\delta$ by training \textsc{InceptionXML-LF} on data generated with varying values of $\delta$. 
As shown in \autoref{tbl:ablation_smooth_labels}, the empirical performance peaks at a $\delta$ value of $0.1$ which is sufficient to suppresses the impact of noisy correlations. 
Higher values of $\delta$ tend to suppress useful information.

\section{Other Related Work} 
\label{sec:related}
Prior works in XMC focused on annotating long-text documents, consisting of hundreds of word tokens, such as those encountered in tagging for Wikipedia \citep{Babbar17, Khandagale19, attentionxml, schultheis2022speeding} with numeric label IDs.
Most recent works under this setting were aimed towards scaling up transformer encoders for the XMC task \citep{zhang2021fast, CascadeXML}. 

\paragraph{\textbf{Exploiting Correlations in XMC}} 
For XMC datasets endowed with label features, there exist three correlations that can be exploited for better representation learning : (i) query-label, (ii) query-query, and (iii) label-label correlations. 
Recent works have been successful in leveraging label features and pushing state-of-the-art by exploiting the first two correlations. 
For example, \textsc{SiameseXML} and \textsc{NGAME} \citep{Siamese, NGAME} employ a two-tower pre-training stage applying contrastive learning between an input text and its corresponding label features. 
\textsc{GalaXC} \citep{GalaXC} \& \textsc{PINA} \citep{chien2023pina}, motivated by graph convolutional networks, create a combined query-label bipartite graph to aggregate predicted instance neighbourhood. This approach, however, leads to a multifold increase in the memory footprint.
\textsc{Decaf} and \textsc{Eclare} \citep{Decaf, Eclare} make architectural additions to embed label-text embeddings (LTE) and graph-augmented label embeddings (GALE) in each label's OVA classifier to exploit higher order correlations from the random walk graph. 
\textsc{PINA}, in its pre-training step, leverages label features as data points, but does so by expanding the label space $\{0, 1\}^L$ to also include instances as $\{0, 1\}^{L+N}$ leveraging the self-annotation property of labels \citep{Siamese} and inverting the initial instance-label mappings to have instances $\instance[i]$ as labels for label features $\labelfeature[l]$ as data points. This, however, leads to an explosion in an already enormous label space. 
In this work, we find that a significant amount of information can be learned by modelling label-label correlations, which existing methods fail to leverage. 

\paragraph{\textbf{Two-tower Models \& Classifier Learning}} Typically, due to the single-annotation nature of most dense retrieval datasets \citep{nguyen2016ms, naturalQ, triviaQA}, two-tower models \citep{karpukhin2020dense} solving this task eliminate classifiers in favour of modelling implicit correlations by bringing query-document embeddings closer in the latent space of the encoders. 
These works are conventionally aimed at improving encoder representations by innovating on hard-negative mining \citep{zhang2021adversarial, xiong2020approximate, lu2022ernie}, teacher-model distillation \citep{qu2020rocketqa, ren2021rocketqav2} and combined dense-sparse training strategies \citep{khattab2020colbert}.
While these approaches result in enhanced encoders, the multilabel nature of XMC makes them, in itself, insufficient for this domain. This has been demonstrated in two-stage XMC works like \citet{Siamese, NGAME, renee} where these frameworks go beyond two-tower training and train classifiers with a frozen encoder in the second stage for better empirical performance. 
While a concurrent work \cite{DSoftmax} does show that dual-encoder XMC models can outperform classifiers, but requires significant computational resources to scale the contrastive loss across the entire label space.

\section{Conclusion}
In this paper, we proposed \textit{Gandalf}, a strategy to learn label correlations, a notoriously difficult challenge. In contrast to previous works which model these correlations implicitly through model training, we propose a supervised approach to explicitly learn them by leveraging the inherent query-label symmetry in short-text extreme classification. We further performed extensive experimentation by implementing on various SOTA XMC methods and demonstrated dramatic increases in prediction performances uniformly across all methods. Moreover, this is achieved with frugal architectures without incurring any computational overheads in inference latency or training memory footprint. We hope our treatment of label correlations in this domain will spur further research towards crafting data-points with more expressive annotations, and further extend it to long-text XMC approaches where the instance-label symmetry is quite ambiguous. 
\bibliographystyle{ACM-Reference-Format}
\balance
\bibliography{main}


\begin{thebibliography}{45}


\ifx \showCODEN    \undefined \def \showCODEN     #1{\unskip}     \fi
\ifx \showDOI      \undefined \def \showDOI       #1{#1}\fi
\ifx \showISBNx    \undefined \def \showISBNx     #1{\unskip}     \fi
\ifx \showISBNxiii \undefined \def \showISBNxiii  #1{\unskip}     \fi
\ifx \showISSN     \undefined \def \showISSN      #1{\unskip}     \fi
\ifx \showLCCN     \undefined \def \showLCCN      #1{\unskip}     \fi
\ifx \shownote     \undefined \def \shownote      #1{#1}          \fi
\ifx \showarticletitle \undefined \def \showarticletitle #1{#1}   \fi
\ifx \showURL      \undefined \def \showURL       {\relax}        \fi
\providecommand\bibfield[2]{#2}
\providecommand\bibinfo[2]{#2}
\providecommand\natexlab[1]{#1}
\providecommand\showeprint[2][]{arXiv:#2}

\bibitem[Adamic and Huberman(2002)]%
        {adamic2002zipf}
\bibfield{author}{\bibinfo{person}{Lada~A Adamic} {and} \bibinfo{person}{Bernardo~A Huberman}.} \bibinfo{year}{2002}\natexlab{}.
\newblock \showarticletitle{Zipf's law and the Internet.}
\newblock \bibinfo{journal}{\emph{Glottometrics}} \bibinfo{volume}{3}, \bibinfo{number}{1} (\bibinfo{year}{2002}), \bibinfo{pages}{143--150}.
\newblock


\bibitem[Anonymous(2024)]%
        {anonymous2024enhancing}
\bibfield{author}{\bibinfo{person}{Anonymous}.} \bibinfo{year}{2024}\natexlab{}.
\newblock \showarticletitle{Enhancing Tail Performance in Extreme Classifiers by Label Variance Reduction}. In \bibinfo{booktitle}{\emph{The Twelfth International Conference on Learning Representations}}.
\newblock
\urldef\tempurl%
\url{https://openreview.net/forum?id=6ARlSgun7J}
\showURL{%
\tempurl}


\bibitem[Babbar and Sch\"{o}lkopf(2017)]%
        {Babbar17}
\bibfield{author}{\bibinfo{person}{R. Babbar} {and} \bibinfo{person}{B. Sch\"{o}lkopf}.} \bibinfo{year}{2017}\natexlab{}.
\newblock \showarticletitle{{DiSMEC: Distributed Sparse Machines for Extreme Multi-label Classification}}. In \bibinfo{booktitle}{\emph{WSDM}}.
\newblock


\bibitem[Babbar and Sch\"{o}lkopf(2019)]%
        {ProXML}
\bibfield{author}{\bibinfo{person}{R. Babbar} {and} \bibinfo{person}{B. Sch\"{o}lkopf}.} \bibinfo{year}{2019}\natexlab{}.
\newblock \showarticletitle{{Data scarcity, robustness and extreme multi-label classification}}.
\newblock \bibinfo{journal}{\emph{Machine Learning}}  \bibinfo{volume}{108} (\bibinfo{year}{2019}), \bibinfo{pages}{1329--1351}.
\newblock


\bibitem[Chiang et~al\mbox{.}(2019)]%
        {clusterGCN}
\bibfield{author}{\bibinfo{person}{Wei-Lin Chiang}, \bibinfo{person}{Xuanqing Liu}, \bibinfo{person}{Si Si}, \bibinfo{person}{Yang Li}, \bibinfo{person}{Samy Bengio}, {and} \bibinfo{person}{Cho-Jui Hsieh}.} \bibinfo{year}{2019}\natexlab{}.
\newblock \showarticletitle{Cluster-GCN: An Efficient Algorithm for Training Deep and Large Graph Convolutional Networks}. In \bibinfo{booktitle}{\emph{Proceedings of the 25th ACM SIGKDD International Conference on Knowledge Discovery \& Data Mining}} (Anchorage, AK, USA) \emph{(\bibinfo{series}{KDD '19})}. \bibinfo{publisher}{Association for Computing Machinery}, \bibinfo{address}{New York, NY, USA}, \bibinfo{pages}{257–266}.
\newblock
\showISBNx{9781450362016}
\urldef\tempurl%
\url{https://doi.org/10.1145/3292500.3330925}
\showDOI{\tempurl}


\bibitem[Chien et~al\mbox{.}(2023)]%
        {chien2023pina}
\bibfield{author}{\bibinfo{person}{Eli Chien}, \bibinfo{person}{Jiong Zhang}, \bibinfo{person}{Cho-Jui Hsieh}, \bibinfo{person}{Jyun-Yu Jiang}, \bibinfo{person}{Wei-Cheng Chang}, \bibinfo{person}{Olgica Milenkovic}, {and} \bibinfo{person}{Hsiang-Fu Yu}.} \bibinfo{year}{2023}\natexlab{}.
\newblock \showarticletitle{PINA: Leveraging Side Information in eXtreme Multi-label Classification via Predicted Instance Neighborhood Aggregation}.
\newblock \bibinfo{journal}{\emph{arXiv preprint arXiv:2305.12349}} (\bibinfo{year}{2023}).
\newblock


\bibitem[Dahiya et~al\mbox{.}(2021a)]%
        {Siamese}
\bibfield{author}{\bibinfo{person}{Kunal Dahiya}, \bibinfo{person}{Ananye Agarwal}, \bibinfo{person}{Deepak Saini}, \bibinfo{person}{Gururaj K}, \bibinfo{person}{Jian Jiao}, \bibinfo{person}{Amit Singh}, \bibinfo{person}{Sumeet Agarwal}, \bibinfo{person}{Purushottam Kar}, {and} \bibinfo{person}{Manik Varma}.} \bibinfo{year}{2021}\natexlab{a}.
\newblock \showarticletitle{SiameseXML: Siamese Networks meet Extreme Classifiers with 100M Labels}. In \bibinfo{booktitle}{\emph{Proceedings of the 38th International Conference on Machine Learning}} \emph{(\bibinfo{series}{Proceedings of Machine Learning Research}, Vol.~\bibinfo{volume}{139})}. \bibinfo{publisher}{PMLR}, \bibinfo{pages}{2330--2340}.
\newblock
\urldef\tempurl%
\url{https://proceedings.mlr.press/v139/dahiya21a.html}
\showURL{%
\tempurl}


\bibitem[Dahiya et~al\mbox{.}(2023)]%
        {NGAME}
\bibfield{author}{\bibinfo{person}{Kunal Dahiya}, \bibinfo{person}{Nilesh Gupta}, \bibinfo{person}{Deepak Saini}, \bibinfo{person}{Akshay Soni}, \bibinfo{person}{Yajun Wang}, \bibinfo{person}{Kushal Dave}, \bibinfo{person}{Jian Jiao}, \bibinfo{person}{Gururaj K}, \bibinfo{person}{Prasenjit Dey}, \bibinfo{person}{Amit Singh}, {et~al\mbox{.}}} \bibinfo{year}{2023}\natexlab{}.
\newblock \showarticletitle{NGAME: Negative Mining-aware Mini-batching for Extreme Classification}. In \bibinfo{booktitle}{\emph{Proceedings of the Sixteenth ACM International Conference on Web Search and Data Mining}}. \bibinfo{pages}{258--266}.
\newblock


\bibitem[Dahiya et~al\mbox{.}(2021b)]%
        {Astec}
\bibfield{author}{\bibinfo{person}{Kunal Dahiya}, \bibinfo{person}{Deepak Saini}, \bibinfo{person}{Anshul Mittal}, \bibinfo{person}{Ankush Shaw}, \bibinfo{person}{Kushal Dave}, \bibinfo{person}{Akshay Soni}, \bibinfo{person}{Himanshu Jain}, \bibinfo{person}{Sumeet Agarwal}, {and} \bibinfo{person}{Manik Varma}.} \bibinfo{year}{2021}\natexlab{b}.
\newblock \showarticletitle{DeepXML: A Deep Extreme Multi-Label Learning Framework Applied to Short Text Documents}. In \bibinfo{booktitle}{\emph{Proceedings of the 14th ACM International Conference on Web Search and Data Mining}} (Virtual Event, Israel) \emph{(\bibinfo{series}{WSDM '21})}. \bibinfo{publisher}{Association for Computing Machinery}, \bibinfo{address}{New York, NY, USA}, \bibinfo{pages}{31–39}.
\newblock
\showISBNx{9781450382977}
\urldef\tempurl%
\url{https://doi.org/10.1145/3437963.3441810}
\showDOI{\tempurl}


\bibitem[Dembczy{\'n}ski et~al\mbox{.}(2012)]%
        {dembczynski2012label}
\bibfield{author}{\bibinfo{person}{Krzysztof Dembczy{\'n}ski}, \bibinfo{person}{Willem Waegeman}, \bibinfo{person}{Weiwei Cheng}, {and} \bibinfo{person}{Eyke H{\"u}llermeier}.} \bibinfo{year}{2012}\natexlab{}.
\newblock \showarticletitle{On label dependence and loss minimization in multi-label classification}.
\newblock \bibinfo{journal}{\emph{Machine Learning}}  \bibinfo{volume}{88} (\bibinfo{year}{2012}), \bibinfo{pages}{5--45}.
\newblock


\bibitem[Guo et~al\mbox{.}(2019)]%
        {Guo2019}
\bibfield{author}{\bibinfo{person}{C. Guo}, \bibinfo{person}{A. Mousavi}, \bibinfo{person}{X. Wu}, \bibinfo{person}{Daniel~N. Holtmann-Rice}, \bibinfo{person}{S. Kale}, \bibinfo{person}{S. Reddi}, {and} \bibinfo{person}{S. Kumar}.} \bibinfo{year}{2019}\natexlab{}.
\newblock \showarticletitle{{Breaking the Glass Ceiling for Embedding-Based Classifiers for Large Output Spaces}}. In \bibinfo{booktitle}{\emph{NeurIPS}}.
\newblock


\bibitem[Gupta et~al\mbox{.}(2023)]%
        {DSoftmax}
\bibfield{author}{\bibinfo{person}{Nilesh Gupta}, \bibinfo{person}{Devvrit Khatri}, \bibinfo{person}{Ankit~S Rawat}, \bibinfo{person}{Srinadh Bhojanapalli}, \bibinfo{person}{Prateek Jain}, {and} \bibinfo{person}{Inderjit~S Dhillon}.} \bibinfo{year}{2023}\natexlab{}.
\newblock \bibinfo{title}{Efficacy of Dual-Encoders for Extreme Multi-Label Classification}.
\newblock
\newblock
\showeprint[arxiv]{2310.10636}~[cs.LG]


\bibitem[Hu et~al\mbox{.}(2020)]%
        {OpenGraphBenchmark}
\bibfield{author}{\bibinfo{person}{Weihua Hu}, \bibinfo{person}{Matthias Fey}, \bibinfo{person}{Marinka Zitnik}, \bibinfo{person}{Yuxiao Dong}, \bibinfo{person}{Hongyu Ren}, \bibinfo{person}{Bowen Liu}, \bibinfo{person}{Michele Catasta}, {and} \bibinfo{person}{Jure Leskovec}.} \bibinfo{year}{2020}\natexlab{}.
\newblock \showarticletitle{Open Graph Benchmark: Datasets for Machine Learning on Graphs}. In \bibinfo{booktitle}{\emph{Proceedings of the 34th International Conference on Neural Information Processing Systems}} (Vancouver, BC, Canada) \emph{(\bibinfo{series}{NIPS'20})}. \bibinfo{publisher}{Curran Associates Inc.}, \bibinfo{address}{Red Hook, NY, USA}, Article \bibinfo{articleno}{1855}, \bibinfo{numpages}{16}~pages.
\newblock
\showISBNx{9781713829546}


\bibitem[H{\"u}llermeier et~al\mbox{.}(2022)]%
        {hullermeier2022flexible}
\bibfield{author}{\bibinfo{person}{Eyke H{\"u}llermeier}, \bibinfo{person}{Marcel Wever}, \bibinfo{person}{Eneldo Loza~Mencia}, \bibinfo{person}{Johannes F{\"u}rnkranz}, {and} \bibinfo{person}{Michael Rapp}.} \bibinfo{year}{2022}\natexlab{}.
\newblock \showarticletitle{A flexible class of dependence-aware multi-label loss functions}.
\newblock \bibinfo{journal}{\emph{Machine Learning}} \bibinfo{volume}{111}, \bibinfo{number}{2} (\bibinfo{year}{2022}), \bibinfo{pages}{713--737}.
\newblock


\bibitem[Jain et~al\mbox{.}(2019)]%
        {Jain2019Slice}
\bibfield{author}{\bibinfo{person}{Himanshu Jain}, \bibinfo{person}{Venkatesh Balasubramanian}, \bibinfo{person}{Bhanu Chunduri}, {and} \bibinfo{person}{Manik Varma}.} \bibinfo{year}{2019}\natexlab{}.
\newblock \showarticletitle{Slice: Scalable Linear Extreme Classifiers Trained on 100 Million Labels for Related Searches}.
\newblock \bibinfo{journal}{\emph{Proceedings of the Twelfth ACM International Conference on Web Search and Data Mining}} (\bibinfo{year}{2019}).
\newblock


\bibitem[Jain et~al\mbox{.}(2016)]%
        {jain2016extreme}
\bibfield{author}{\bibinfo{person}{Himanshu Jain}, \bibinfo{person}{Yashoteja Prabhu}, {and} \bibinfo{person}{Manik Varma}.} \bibinfo{year}{2016}\natexlab{}.
\newblock \showarticletitle{Extreme multi-label loss functions for recommendation, tagging, ranking \& other missing label applications}. In \bibinfo{booktitle}{\emph{KDD}}. \bibinfo{pages}{935--944}.
\newblock


\bibitem[Jain et~al\mbox{.}(2023)]%
        {renee}
\bibfield{author}{\bibinfo{person}{Vidit Jain}, \bibinfo{person}{Jatin Prakash}, \bibinfo{person}{Deepak Saini}, \bibinfo{person}{Jian Jiao}, \bibinfo{person}{Ramachandran Ramjee}, {and} \bibinfo{person}{Manik Varma}.} \bibinfo{year}{2023}\natexlab{}.
\newblock \showarticletitle{Renee: End-to-end training of extreme classification models}.
\newblock \bibinfo{journal}{\emph{Proceedings of Machine Learning and Systems}} (\bibinfo{year}{2023}).
\newblock


\bibitem[Joshi et~al\mbox{.}(2017)]%
        {triviaQA}
\bibfield{author}{\bibinfo{person}{Mandar Joshi}, \bibinfo{person}{Eunsol Choi}, \bibinfo{person}{Daniel Weld}, {and} \bibinfo{person}{Luke Zettlemoyer}.} \bibinfo{year}{2017}\natexlab{}.
\newblock \showarticletitle{{T}rivia{QA}: A Large Scale Distantly Supervised Challenge Dataset for Reading Comprehension}. In \bibinfo{booktitle}{\emph{Proceedings of the 55th Annual Meeting of the Association for Computational Linguistics (Volume 1: Long Papers)}}. \bibinfo{publisher}{Association for Computational Linguistics}, \bibinfo{address}{Vancouver, Canada}, \bibinfo{pages}{1601--1611}.
\newblock
\urldef\tempurl%
\url{https://doi.org/10.18653/v1/P17-1147}
\showDOI{\tempurl}


\bibitem[Karpukhin et~al\mbox{.}(2020)]%
        {karpukhin2020dense}
\bibfield{author}{\bibinfo{person}{Vladimir Karpukhin}, \bibinfo{person}{Barlas O{\u{g}}uz}, \bibinfo{person}{Sewon Min}, \bibinfo{person}{Patrick Lewis}, \bibinfo{person}{Ledell Wu}, \bibinfo{person}{Sergey Edunov}, \bibinfo{person}{Danqi Chen}, {and} \bibinfo{person}{Wen-tau Yih}.} \bibinfo{year}{2020}\natexlab{}.
\newblock \showarticletitle{Dense passage retrieval for open-domain question answering}.
\newblock \bibinfo{journal}{\emph{arXiv preprint arXiv:2004.04906}} (\bibinfo{year}{2020}).
\newblock


\bibitem[Khandagale et~al\mbox{.}(2020)]%
        {Khandagale19}
\bibfield{author}{\bibinfo{person}{S. Khandagale}, \bibinfo{person}{H. Xiao}, {and} \bibinfo{person}{R. Babbar}.} \bibinfo{year}{2020}\natexlab{}.
\newblock \showarticletitle{{Bonsai: diverse and shallow trees for extreme multi-label classification}}.
\newblock \bibinfo{journal}{\emph{Machine Learning}} \bibinfo{volume}{109}, \bibinfo{number}{11} (\bibinfo{year}{2020}), \bibinfo{pages}{2099--2119}.
\newblock


\bibitem[Kharbanda et~al\mbox{.}(2023)]%
        {kharbanda2021embedding}
\bibfield{author}{\bibinfo{person}{Siddhant Kharbanda}, \bibinfo{person}{Atmadeep Banerjee}, \bibinfo{person}{Devaansh Gupta}, \bibinfo{person}{Akash Palrecha}, {and} \bibinfo{person}{Rohit Babbar}.} \bibinfo{year}{2023}\natexlab{}.
\newblock \showarticletitle{InceptionXML: A Lightweight Framework with Synchronized Negative Sampling for Short Text Extreme Classification}. In \bibinfo{booktitle}{\emph{Proceedings of the 46th International ACM SIGIR Conference on Research and Development in Information Retrieval}} (Taipei, Taiwan) \emph{(\bibinfo{series}{SIGIR '23})}. \bibinfo{publisher}{Association for Computing Machinery}, \bibinfo{address}{Taipei, Taiwan}, \bibinfo{pages}{760–769}.
\newblock
\showISBNx{9781450394086}
\urldef\tempurl%
\url{https://doi.org/10.1145/3539618.3591699}
\showDOI{\tempurl}


\bibitem[Kharbanda et~al\mbox{.}(2022)]%
        {CascadeXML}
\bibfield{author}{\bibinfo{person}{Siddhant Kharbanda}, \bibinfo{person}{Atmadeep Banerjee}, \bibinfo{person}{Erik Schultheis}, {and} \bibinfo{person}{Rohit Babbar}.} \bibinfo{year}{2022}\natexlab{}.
\newblock \showarticletitle{CascadeXML: Rethinking Transformers for End-to-end Multi-resolution Training in Extreme Multi-label Classification}. In \bibinfo{booktitle}{\emph{Advances in Neural Information Processing Systems}}, Vol.~\bibinfo{volume}{35}. \bibinfo{publisher}{Curran Associates, Inc.}, \bibinfo{pages}{2074--2087}.
\newblock
\urldef\tempurl%
\url{https://proceedings.neurips.cc/paper_files/paper/2022/file/0e0157ce5ea15831072be4744cbd5334-Paper-Conference.pdf}
\showURL{%
\tempurl}


\bibitem[Khattab and Zaharia(2020)]%
        {khattab2020colbert}
\bibfield{author}{\bibinfo{person}{Omar Khattab} {and} \bibinfo{person}{Matei Zaharia}.} \bibinfo{year}{2020}\natexlab{}.
\newblock \showarticletitle{Colbert: Efficient and effective passage search via contextualized late interaction over bert}. In \bibinfo{booktitle}{\emph{Proceedings of the 43rd International ACM SIGIR conference on research and development in Information Retrieval}}. \bibinfo{pages}{39--48}.
\newblock


\bibitem[Kwiatkowski et~al\mbox{.}(2019)]%
        {naturalQ}
\bibfield{author}{\bibinfo{person}{Tom Kwiatkowski}, \bibinfo{person}{Jennimaria Palomaki}, \bibinfo{person}{Olivia Redfield}, \bibinfo{person}{Michael Collins}, \bibinfo{person}{Ankur Parikh}, \bibinfo{person}{Chris Alberti}, \bibinfo{person}{Danielle Epstein}, \bibinfo{person}{Illia Polosukhin}, \bibinfo{person}{Jacob Devlin}, \bibinfo{person}{Kenton Lee}, \bibinfo{person}{Kristina Toutanova}, \bibinfo{person}{Llion Jones}, \bibinfo{person}{Matthew Kelcey}, \bibinfo{person}{Ming-Wei Chang}, \bibinfo{person}{Andrew~M. Dai}, \bibinfo{person}{Jakob Uszkoreit}, \bibinfo{person}{Quoc Le}, {and} \bibinfo{person}{Slav Petrov}.} \bibinfo{year}{2019}\natexlab{}.
\newblock \showarticletitle{Natural Questions: A Benchmark for Question Answering Research}.
\newblock \bibinfo{journal}{\emph{Transactions of the Association for Computational Linguistics}}  \bibinfo{volume}{7} (\bibinfo{year}{2019}), \bibinfo{pages}{452--466}.
\newblock
\urldef\tempurl%
\url{https://doi.org/10.1162/tacl_a_00276}
\showDOI{\tempurl}


\bibitem[Lu et~al\mbox{.}(2022)]%
        {lu2022ernie}
\bibfield{author}{\bibinfo{person}{Yuxiang Lu}, \bibinfo{person}{Yiding Liu}, \bibinfo{person}{Jiaxiang Liu}, \bibinfo{person}{Yunsheng Shi}, \bibinfo{person}{Zhengjie Huang}, \bibinfo{person}{Shikun Feng~Yu Sun}, \bibinfo{person}{Hao Tian}, \bibinfo{person}{Hua Wu}, \bibinfo{person}{Shuaiqiang Wang}, \bibinfo{person}{Dawei Yin}, {et~al\mbox{.}}} \bibinfo{year}{2022}\natexlab{}.
\newblock \showarticletitle{Ernie-search: Bridging cross-encoder with dual-encoder via self on-the-fly distillation for dense passage retrieval}.
\newblock \bibinfo{journal}{\emph{arXiv preprint arXiv:2205.09153}} (\bibinfo{year}{2022}).
\newblock


\bibitem[Menon et~al\mbox{.}(2019)]%
        {menon2019multilabel}
\bibfield{author}{\bibinfo{person}{Aditya~K Menon}, \bibinfo{person}{Ankit~Singh Rawat}, \bibinfo{person}{Sashank Reddi}, {and} \bibinfo{person}{Sanjiv Kumar}.} \bibinfo{year}{2019}\natexlab{}.
\newblock \showarticletitle{Multilabel reductions: what is my loss optimising?}. In \bibinfo{booktitle}{\emph{Advances in Neural Information Processing Systems}}, \bibfield{editor}{\bibinfo{person}{H.~Wallach}, \bibinfo{person}{H.~Larochelle}, \bibinfo{person}{A.~Beygelzimer}, \bibinfo{person}{F.~d\textquotesingle Alch\'{e}-Buc}, \bibinfo{person}{E.~Fox}, {and} \bibinfo{person}{R.~Garnett}} (Eds.), Vol.~\bibinfo{volume}{32}. \bibinfo{publisher}{Curran Associates, Inc.}
\newblock
\urldef\tempurl%
\url{https://proceedings.neurips.cc/paper_files/paper/2019/file/da647c549dde572c2c5edc4f5bef039c-Paper.pdf}
\showURL{%
\tempurl}


\bibitem[Mittal et~al\mbox{.}(2021a)]%
        {Decaf}
\bibfield{author}{\bibinfo{person}{Anshul Mittal}, \bibinfo{person}{Kunal Dahiya}, \bibinfo{person}{Sheshansh Agrawal}, \bibinfo{person}{Deepak Saini}, \bibinfo{person}{Sumeet Agarwal}, \bibinfo{person}{Purushottam Kar}, {and} \bibinfo{person}{Manik Varma}.} \bibinfo{year}{2021}\natexlab{a}.
\newblock \showarticletitle{DECAF: Deep Extreme Classification with Label Features}. In \bibinfo{booktitle}{\emph{Proceedings of the 14th ACM International Conference on Web Search and Data Mining}} (Virtual Event, Israel) \emph{(\bibinfo{series}{WSDM '21})}. \bibinfo{publisher}{Association for Computing Machinery}, \bibinfo{address}{New York, NY, USA}, \bibinfo{pages}{49–57}.
\newblock
\showISBNx{9781450382977}
\urldef\tempurl%
\url{https://doi.org/10.1145/3437963.3441807}
\showDOI{\tempurl}


\bibitem[Mittal et~al\mbox{.}(2021b)]%
        {Eclare}
\bibfield{author}{\bibinfo{person}{Anshul Mittal}, \bibinfo{person}{Noveen Sachdeva}, \bibinfo{person}{Sheshansh Agrawal}, \bibinfo{person}{Sumeet Agarwal}, \bibinfo{person}{Purushottam Kar}, {and} \bibinfo{person}{Manik Varma}.} \bibinfo{year}{2021}\natexlab{b}.
\newblock \showarticletitle{ECLARE: Extreme Classification with Label Graph Correlations}. In \bibinfo{booktitle}{\emph{Proceedings of the Web Conference 2021}} (Ljubljana, Slovenia) \emph{(\bibinfo{series}{WWW '21})}. \bibinfo{publisher}{Association for Computing Machinery}, \bibinfo{address}{New York, NY, USA}, \bibinfo{pages}{3721–3732}.
\newblock
\showISBNx{9781450383127}
\urldef\tempurl%
\url{https://doi.org/10.1145/3442381.3449815}
\showDOI{\tempurl}


\bibitem[Nguyen et~al\mbox{.}(2016)]%
        {nguyen2016ms}
\bibfield{author}{\bibinfo{person}{Tri Nguyen}, \bibinfo{person}{Mir Rosenberg}, \bibinfo{person}{Xia Song}, \bibinfo{person}{Jianfeng Gao}, \bibinfo{person}{Saurabh Tiwary}, \bibinfo{person}{Rangan Majumder}, {and} \bibinfo{person}{Li Deng}.} \bibinfo{year}{2016}\natexlab{}.
\newblock \showarticletitle{Ms marco: A human-generated machine reading comprehension dataset}.
\newblock  (\bibinfo{year}{2016}).
\newblock


\bibitem[Partalas et~al\mbox{.}(2015)]%
        {partalas2015lshtc}
\bibfield{author}{\bibinfo{person}{Ioannis Partalas}, \bibinfo{person}{Aris Kosmopoulos}, \bibinfo{person}{Nicolas Baskiotis}, \bibinfo{person}{Thierry Artieres}, \bibinfo{person}{George Paliouras}, \bibinfo{person}{Eric Gaussier}, \bibinfo{person}{Ion Androutsopoulos}, \bibinfo{person}{Massih-Reza Amini}, {and} \bibinfo{person}{Patrick Galinari}.} \bibinfo{year}{2015}\natexlab{}.
\newblock \showarticletitle{Lshtc: A benchmark for large-scale text classification}.
\newblock \bibinfo{journal}{\emph{arXiv preprint arXiv:1503.08581}} (\bibinfo{year}{2015}).
\newblock


\bibitem[Prabhu et~al\mbox{.}(2018)]%
        {Parabel}
\bibfield{author}{\bibinfo{person}{Yashoteja Prabhu}, \bibinfo{person}{Anil Kag}, \bibinfo{person}{Shrutendra Harsola}, \bibinfo{person}{Rahul Agrawal}, {and} \bibinfo{person}{Manik Varma}.} \bibinfo{year}{2018}\natexlab{}.
\newblock \showarticletitle{Parabel: Partitioned Label Trees for Extreme Classification with Application to Dynamic Search Advertising}. In \bibinfo{booktitle}{\emph{Proceedings of the 2018 World Wide Web Conference}} (Lyon, France) \emph{(\bibinfo{series}{WWW '18})}. \bibinfo{publisher}{International World Wide Web Conferences Steering Committee}, \bibinfo{address}{Republic and Canton of Geneva, CHE}, \bibinfo{pages}{993–1002}.
\newblock
\showISBNx{9781450356398}
\urldef\tempurl%
\url{https://doi.org/10.1145/3178876.3185998}
\showDOI{\tempurl}


\bibitem[Qaraei et~al\mbox{.}(2021)]%
        {qaraei2021convex}
\bibfield{author}{\bibinfo{person}{Mohammadreza Qaraei}, \bibinfo{person}{Erik Schultheis}, \bibinfo{person}{Priyanshu Gupta}, {and} \bibinfo{person}{Rohit Babbar}.} \bibinfo{year}{2021}\natexlab{}.
\newblock \showarticletitle{Convex Surrogates for Unbiased Loss Functions in Extreme Classification With Missing Labels}. In \bibinfo{booktitle}{\emph{Proceedings of the Web Conference 2021}}. \bibinfo{pages}{3711--3720}.
\newblock


\bibitem[Qu et~al\mbox{.}(2021)]%
        {qu2020rocketqa}
\bibfield{author}{\bibinfo{person}{Yingqi Qu}, \bibinfo{person}{Yuchen Ding}, \bibinfo{person}{Jing Liu}, \bibinfo{person}{Kai Liu}, \bibinfo{person}{Ruiyang Ren}, \bibinfo{person}{Wayne~Xin Zhao}, \bibinfo{person}{Daxiang Dong}, \bibinfo{person}{Hua Wu}, {and} \bibinfo{person}{Haifeng Wang}.} \bibinfo{year}{2021}\natexlab{}.
\newblock \showarticletitle{{R}ocket{QA}: An Optimized Training Approach to Dense Passage Retrieval for Open-Domain Question Answering}. In \bibinfo{booktitle}{\emph{Proceedings of the 2021 Conference of the North American Chapter of the Association for Computational Linguistics: Human Language Technologies}}. \bibinfo{publisher}{Association for Computational Linguistics}, \bibinfo{address}{Online}, \bibinfo{pages}{5835--5847}.
\newblock
\urldef\tempurl%
\url{https://doi.org/10.18653/v1/2021.naacl-main.466}
\showDOI{\tempurl}


\bibitem[Ren et~al\mbox{.}(2021)]%
        {ren2021rocketqav2}
\bibfield{author}{\bibinfo{person}{Ruiyang Ren}, \bibinfo{person}{Yingqi Qu}, \bibinfo{person}{Jing Liu}, \bibinfo{person}{Wayne~Xin Zhao}, \bibinfo{person}{QiaoQiao She}, \bibinfo{person}{Hua Wu}, \bibinfo{person}{Haifeng Wang}, {and} \bibinfo{person}{Ji-Rong Wen}.} \bibinfo{year}{2021}\natexlab{}.
\newblock \showarticletitle{{R}ocket{QA}v2: A Joint Training Method for Dense Passage Retrieval and Passage Re-ranking}. In \bibinfo{booktitle}{\emph{Proceedings of the 2021 Conference on Empirical Methods in Natural Language Processing}}. \bibinfo{publisher}{Association for Computational Linguistics}, \bibinfo{address}{Online and Punta Cana, Dominican Republic}, \bibinfo{pages}{2825--2835}.
\newblock
\urldef\tempurl%
\url{https://doi.org/10.18653/v1/2021.emnlp-main.224}
\showDOI{\tempurl}


\bibitem[Saini et~al\mbox{.}(2021)]%
        {GalaXC}
\bibfield{author}{\bibinfo{person}{Deepak Saini}, \bibinfo{person}{Arnav~Kumar Jain}, \bibinfo{person}{Kushal Dave}, \bibinfo{person}{Jian Jiao}, \bibinfo{person}{Amit Singh}, \bibinfo{person}{Ruofei Zhang}, {and} \bibinfo{person}{Manik Varma}.} \bibinfo{year}{2021}\natexlab{}.
\newblock \showarticletitle{GalaXC: Graph neural networks with labelwise attention for extreme classification}. In \bibinfo{booktitle}{\emph{ACM International World Wide Web Conference}}.
\newblock
\urldef\tempurl%
\url{https://www.microsoft.com/en-us/research/publication/galaxc/}
\showURL{%
\tempurl}


\bibitem[Schultheis and Babbar(2022)]%
        {schultheis2022speeding}
\bibfield{author}{\bibinfo{person}{Erik Schultheis} {and} \bibinfo{person}{Rohit Babbar}.} \bibinfo{year}{2022}\natexlab{}.
\newblock \showarticletitle{Speeding-up one-versus-all training for extreme classification via mean-separating initialization}.
\newblock \bibinfo{journal}{\emph{Machine Learning}} \bibinfo{volume}{111}, \bibinfo{number}{11} (\bibinfo{year}{2022}), \bibinfo{pages}{3953--3976}.
\newblock


\bibitem[Schultheis et~al\mbox{.}(2022)]%
        {schultheis2022missing}
\bibfield{author}{\bibinfo{person}{Erik Schultheis}, \bibinfo{person}{Marek Wydmuch}, \bibinfo{person}{Rohit Babbar}, {and} \bibinfo{person}{Krzysztof Dembczynski}.} \bibinfo{year}{2022}\natexlab{}.
\newblock \showarticletitle{On missing labels, long-tails and propensities in extreme multi-label classification}. In \bibinfo{booktitle}{\emph{Proceedings of the 28th ACM SIGKDD Conference on Knowledge Discovery and Data Mining}}. \bibinfo{pages}{1547--1557}.
\newblock


\bibitem[Schultheis et~al\mbox{.}(2024)]%
        {schultheis2024generalized}
\bibfield{author}{\bibinfo{person}{Erik Schultheis}, \bibinfo{person}{Marek Wydmuch}, \bibinfo{person}{Wojciech Kotlowski}, \bibinfo{person}{Rohit Babbar}, {and} \bibinfo{person}{Krzysztof Dembczynski}.} \bibinfo{year}{2024}\natexlab{}.
\newblock \showarticletitle{Generalized test utilities for long-tail performance in extreme multi-label classification}.
\newblock \bibinfo{journal}{\emph{Advances in Neural Information Processing Systems}}  \bibinfo{volume}{36} (\bibinfo{year}{2024}).
\newblock


\bibitem[Tsatsaronis et~al\mbox{.}(2015)]%
        {tsatsaronis2015overview}
\bibfield{author}{\bibinfo{person}{George Tsatsaronis}, \bibinfo{person}{Georgios Balikas}, \bibinfo{person}{Prodromos Malakasiotis}, \bibinfo{person}{Ioannis Partalas}, \bibinfo{person}{Matthias Zschunke}, \bibinfo{person}{Michael~R Alvers}, \bibinfo{person}{Dirk Weissenborn}, \bibinfo{person}{Anastasia Krithara}, \bibinfo{person}{Sergios Petridis}, \bibinfo{person}{Dimitris Polychronopoulos}, {et~al\mbox{.}}} \bibinfo{year}{2015}\natexlab{}.
\newblock \showarticletitle{An overview of the BIOASQ large-scale biomedical semantic indexing and question answering competition}.
\newblock \bibinfo{journal}{\emph{BMC bioinformatics}} \bibinfo{volume}{16}, \bibinfo{number}{1} (\bibinfo{year}{2015}), \bibinfo{pages}{1--28}.
\newblock


\bibitem[Wydmuch et~al\mbox{.}(2018)]%
        {Wydmuch18}
\bibfield{author}{\bibinfo{person}{M. Wydmuch}, \bibinfo{person}{K. Jasinska}, \bibinfo{person}{M. Kuznetsov}, \bibinfo{person}{R. Busa-Fekete}, {and} \bibinfo{person}{K. Dembczynski}.} \bibinfo{year}{2018}\natexlab{}.
\newblock \showarticletitle{{A no-regret generalization of hierarchical softmax to extreme multi-label classification}}. In \bibinfo{booktitle}{\emph{NIPS}}.
\newblock


\bibitem[Xiong et~al\mbox{.}(2021)]%
        {xiong2020approximate}
\bibfield{author}{\bibinfo{person}{Lee Xiong}, \bibinfo{person}{Chenyan Xiong}, \bibinfo{person}{Ye Li}, \bibinfo{person}{Kwok-Fung Tang}, \bibinfo{person}{Jialin Liu}, \bibinfo{person}{Paul~N. Bennett}, \bibinfo{person}{Junaid Ahmed}, {and} \bibinfo{person}{Arnold Overwijk}.} \bibinfo{year}{2021}\natexlab{}.
\newblock \showarticletitle{Approximate Nearest Neighbor Negative Contrastive Learning for Dense Text Retrieval}. In \bibinfo{booktitle}{\emph{International Conference on Learning Representations}}.
\newblock
\urldef\tempurl%
\url{https://openreview.net/forum?id=zeFrfgyZln}
\showURL{%
\tempurl}


\bibitem[Ye et~al\mbox{.}(2020)]%
        {APLC_XLNet}
\bibfield{author}{\bibinfo{person}{H. Ye}, \bibinfo{person}{Z. Chen}, \bibinfo{person}{D.-H. Wang}, {and} \bibinfo{person}{Davison~B. D.}} \bibinfo{year}{2020}\natexlab{}.
\newblock \showarticletitle{{Pretrained Generalized Autoregressive Model with Adaptive Probabilistic Label Clusters for Extreme Multi-label Text Classification}}. In \bibinfo{booktitle}{\emph{ICML}}.
\newblock


\bibitem[You et~al\mbox{.}(2019)]%
        {attentionxml}
\bibfield{author}{\bibinfo{person}{R. You}, \bibinfo{person}{Z. Zhang}, \bibinfo{person}{Z. Wang}, \bibinfo{person}{S. Dai}, \bibinfo{person}{H. Mamitsuka}, {and} \bibinfo{person}{S. Zhu}.} \bibinfo{year}{2019}\natexlab{}.
\newblock \showarticletitle{{Attentionxml: Label tree-based attention-aware deep model for high-performance extreme multi-label text classification}}. In \bibinfo{booktitle}{\emph{NeurIPS}}.
\newblock


\bibitem[Zhang et~al\mbox{.}(2021b)]%
        {zhang2021adversarial}
\bibfield{author}{\bibinfo{person}{Hang Zhang}, \bibinfo{person}{Yeyun Gong}, \bibinfo{person}{Yelong Shen}, \bibinfo{person}{Jiancheng Lv}, \bibinfo{person}{Nan Duan}, {and} \bibinfo{person}{Weizhu Chen}.} \bibinfo{year}{2021}\natexlab{b}.
\newblock \showarticletitle{Adversarial retriever-ranker for dense text retrieval}.
\newblock \bibinfo{journal}{\emph{arXiv preprint arXiv:2110.03611}} (\bibinfo{year}{2021}).
\newblock


\bibitem[Zhang et~al\mbox{.}(2021a)]%
        {zhang2021fast}
\bibfield{author}{\bibinfo{person}{Jiong Zhang}, \bibinfo{person}{Wei-Cheng Chang}, \bibinfo{person}{Hsiang-Fu Yu}, {and} \bibinfo{person}{Inderjit~S Dhillon}.} \bibinfo{year}{2021}\natexlab{a}.
\newblock \showarticletitle{Fast Multi-Resolution Transformer Fine-tuning for Extreme Multi-label Text Classification}. In \bibinfo{booktitle}{\emph{Advances in Neural Information Processing Systems}}.
\newblock
\urldef\tempurl%
\url{https://openreview.net/forum?id=gjBz22V93a}
\showURL{%
\tempurl}


\end{thebibliography}
\newpage
\appendix
\section{Additional Visualizations}
\label{visualisations}
\paragraph{\textbf{Additional Qualitative Examples}} The same observations in the examples provided in the main paper can be made for these additional examples as well (\autoref{tbl:main_comparisons}). Where initially we mispredict all the labels, with \textsc{Gandalf}, we can correctly predict all the labels.\\

\begin{figure*}[!hb]
    \centering
    \includegraphics[width=0.7\textwidth]{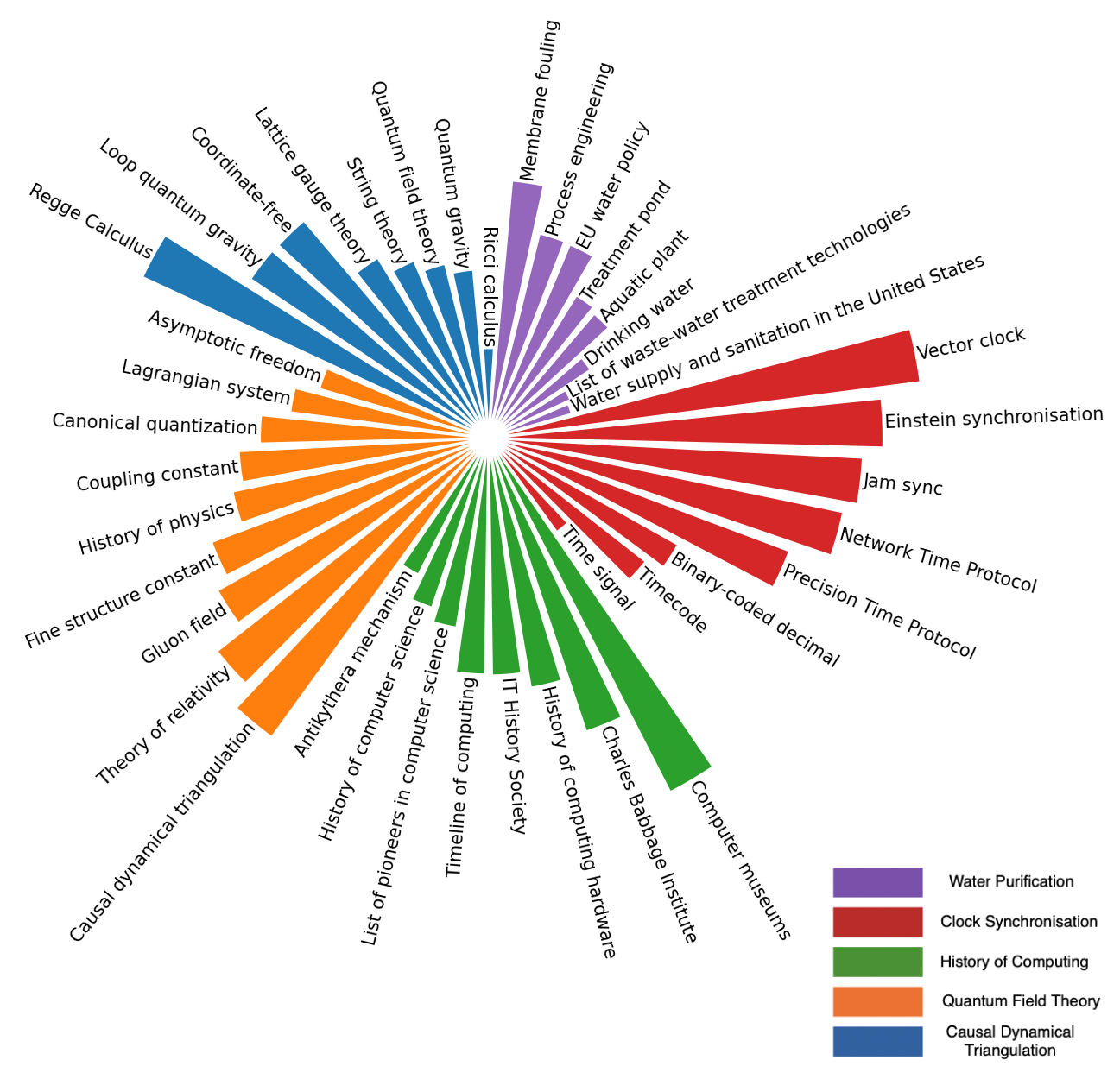}
    \caption{Correlations between labels and their first-order neighbours, as found by the label co-occurrence on the LF-WikiTitles-500K dataset. The legend shows the label in question, the bar chart shows the degree of correlation with its neighbouring labels. Correlated labels often share tokens with each other and/or may be used in the same context.}
    \label{fig:LCG_visual}
\end{figure*}

\paragraph{\textbf{Similarity between Labels and their Annotations}} Each label and their annotations, as discovered by the co-occurrence graph, are semantically similar; in that they share tokens with one another and can be used in the same context. We show this pictorially by plotting the labels and their annotations in order of their co-occurrence in \autoref{fig:LCG_visual}. \\

\begin{figure}[!ht]
  \centering
  \includegraphics[trim={0 0 0 1cm}, clip, width=\columnwidth]{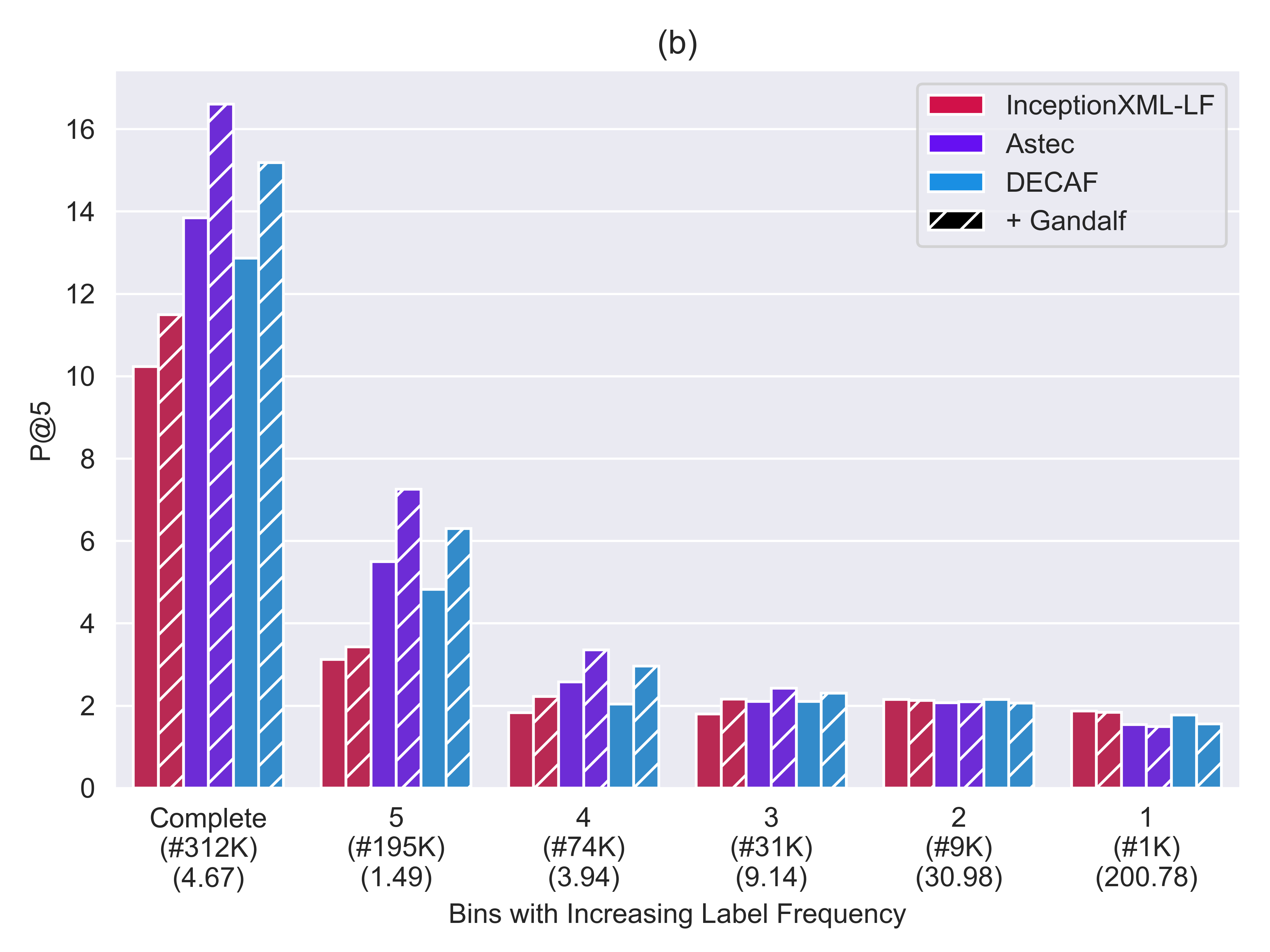}
\caption{Contributions to P@5 in LF-WikiSeeAlsoTitles-320K. The number of labels in each bin is provided after the \# in the second row of the tags on the x-axis. The bottomost row denotes the mean label frequency in that bin. Specifically, note the improvements on tail labels in the earlier bins (5 - 3).}
\label{fig: freq_dist}
\end{figure}

\paragraph{\textbf{Additional Quantile Analysis for Tail Labels}} We show an additional quantile analysis on the LF-WikiSeeAlsoTitles-320K dataset in \autoref{fig: freq_dist}, to demonstrate improvements in labels with different frequencies. The observations remain consistent with that of LF-AmazonTitles-131K in the main paper.\\

\section{Additional Experiments}
\paragraph{\textbf{Coverage Results}} Coverage is an important metric in XMC as it demonstrates the ability of the model to predict tail labels effectively. We provide coverage results on InceptionXML in \autoref{tbl:coverage}, demonstrating that Gandalf learns to predict labels which were previously not being predicted at all. This phenomenon can also be seen in the qualitative results \autoref{tbl:main_comparisons}

\paragraph{\textbf{Comparison against conventional data augmentation strategies}} We compare \textsc{Gandalf} with with existing data augmentation techniques in \autoref{tbl:data_aug}. While no such techniques exist specifically for XMC, we use three baselines: synonym replacement(randomly replacing words in the input text with their synonyms, chosen via BERT similarity), MixUp and Label-MixUp. While the first two are standard data augmentations in NLP, Label-Mixup is a modified version of MixUp that combines the feature of a label feature and input datapoint, which is more suitable for XMC. Notably, Gandalf outperforms all of them with a significant margin:
\begin{table}[!h]
    \begin{adjustbox}{width=\columnwidth,center}
    \begin{tabular}{c|ccc|ccc}
        \toprule
        \textbf{Method} & \textbf{C@1}   & \textbf{C@3}   & \textbf{C@5}   & \textbf{C@1}   & \textbf{C@3}   & \textbf{C@5} \\
        \specialrule{0.70pt}{0.4ex}{0.65ex}
        & \multicolumn{3}{c|}{\textbf{LF-AmazonTitles-131K}} & \multicolumn{3}{c}{\textbf{LF-WikiSeeAlsoTitles-320K}}\\
        \midrule
        InceptionXML & 22.33 & 39.98 & 46.29 & 7.54 & 15.11 & 18.93 \\
        + Gandalf & \textbf{31.04} & \textbf{51.63} & \textbf{58.03} & \textbf{13.28} & \textbf{26.01} & \textbf{32.21}\\
        \bottomrule
    \end{tabular}
    \end{adjustbox}
    \caption{Coverage Results on InceptionXML and \textsc{Gandalf}. Best scores are in \textbf{bold}.}
    \label{tbl:coverage}
\end{table}

\begin{table}[!h]
    \begin{adjustbox}{width=\columnwidth,center}
    \begin{tabular}{c|ccc|ccc}
        \toprule
        \textbf{Method} & \textbf{P@1}   & \textbf{P@3}   & \textbf{P@5}   & \textbf{PSP@1} & \textbf{PSP@3} & \textbf{PSP@5} \\
        \specialrule{0.70pt}{0.4ex}{0.65ex}
        & \multicolumn{6}{c}{\textbf{LF-AmazonTitles-131K}}\\
        \midrule
        InceptionXML & 35.62 & 24.13 & 17.35 & 27.53 & 33.06 & 37.50 \\
        + Synonym Replacement & 35.07 & 23.71 & 17.08 & 27.20 & 32.41 & 36.77 \\
        + MixUp & 35.63 & 24.15 & 17.37 & 27.55 & 33.00 & 37.63 \\
        + Label-MixUp & 37.25 & 25.02 & 17.98 & 29.25 & 34.58 & 39.09 \\
        + Gandalf & \textbf{43.52} & \textbf{29.23} & \textbf{20.92} & \textbf{36.96} & \textbf{42.71} & \textbf{47.64} \\
        \midrule
        & \multicolumn{6}{c}{\textbf{LF-WikiSeeAlsoTitles-320K}}\\
        \midrule
        InceptionXML & 21.53 & 14.19 & 10.66 & 13.06 & 14.87 & 16.33 \\
        + Synonym Replacement & 20.08 & 13.13 & 9.92 & 12.00 & 13.50 & 14.90 \\
        + MixUp & 21.62 & 14.15 & 10.65 & 13.13 & 14.99 & 16.36 \\
        + Label-MixUp & 23.90 & 16.10 & 12.28 & 15.20 & 17.60 & 19.56 \\
        + Gandalf & \textbf{31.31} & \textbf{21.38} & \textbf{16.22} & \textbf{24.31} & \textbf{26.79} & \textbf{28.83} \\
        \bottomrule
    \end{tabular}
    \end{adjustbox}
    \caption{Comparison of conventional data augmentation strategies with the proposed \textsc{Gandalf} method. Best scores are in \textbf{bold}.}
    \label{tbl:data_aug}
\end{table}
\end{document}